\pdfoutput=1

\documentclass[11pt]{article}

\usepackage{ACL2023}
\usepackage{times}

\usepackage[T1]{fontenc}

\usepackage[utf8]{inputenc}

\usepackage{microtype}

\usepackage{graphicx}
\usepackage{amsmath}
\usepackage{amssymb}
\usepackage{booktabs}
\usepackage{multirow}
\usepackage{tabularx}
\usepackage{amsfonts}

\usepackage{color,colortbl}
\usepackage{subfig}
\usepackage{adjustbox}

\usepackage{inconsolata}

\usepackage{caption}
\newcommand{\acro}[0]{ArK}

\newcommand\blfootnote[1]{%
  \begingroup
  \renewcommand\thefootnote{}\footnote{#1}%
  \addtocounter{footnote}{-1}%
  \endgroup
}

\title{ArK: Augmented Reality with Knowledge Interactive Emergent Ability}

\author{
Qiuyuan Huang$^{\ddagger*}$ \And Jae Sung Park$^{\mathsection\ddagger*}$
        \And Abhinav Gupta$^{\dagger\ddagger*}$
        \And Paul Bennett$^{\ddagger}$ \AND Ran Gong$^{\natural}$ \And Subhojit Som$^{\ddagger}$ \And Baolin Peng$^{\ddagger}$ \And Owais Khan Mohammed$^{\ddagger}$ \AND 
        Chris Pal$^{\dagger}$ \And Yejin Choi$^{\mathsection}$ \And Jianfeng Gao$^{\ddagger}$ \AND
  \normalfont$^{\ddagger}${Microsoft Research, Redmond} \And \normalfont$^{\dagger}$ MILA \And  \normalfont$^{\mathsection}${University of Washington} \And \normalfont$^{\natural}$ {UCLA}
  \vspace{-0.3cm}
}

\begin{document}


\twocolumn[{
\renewcommand\twocolumn[1][]{#1}
\maketitle
\begin{center}
    \centering
    \vspace{-15pt}
    \captionsetup{type=figure}
    \includegraphics[width=0.9\textwidth]{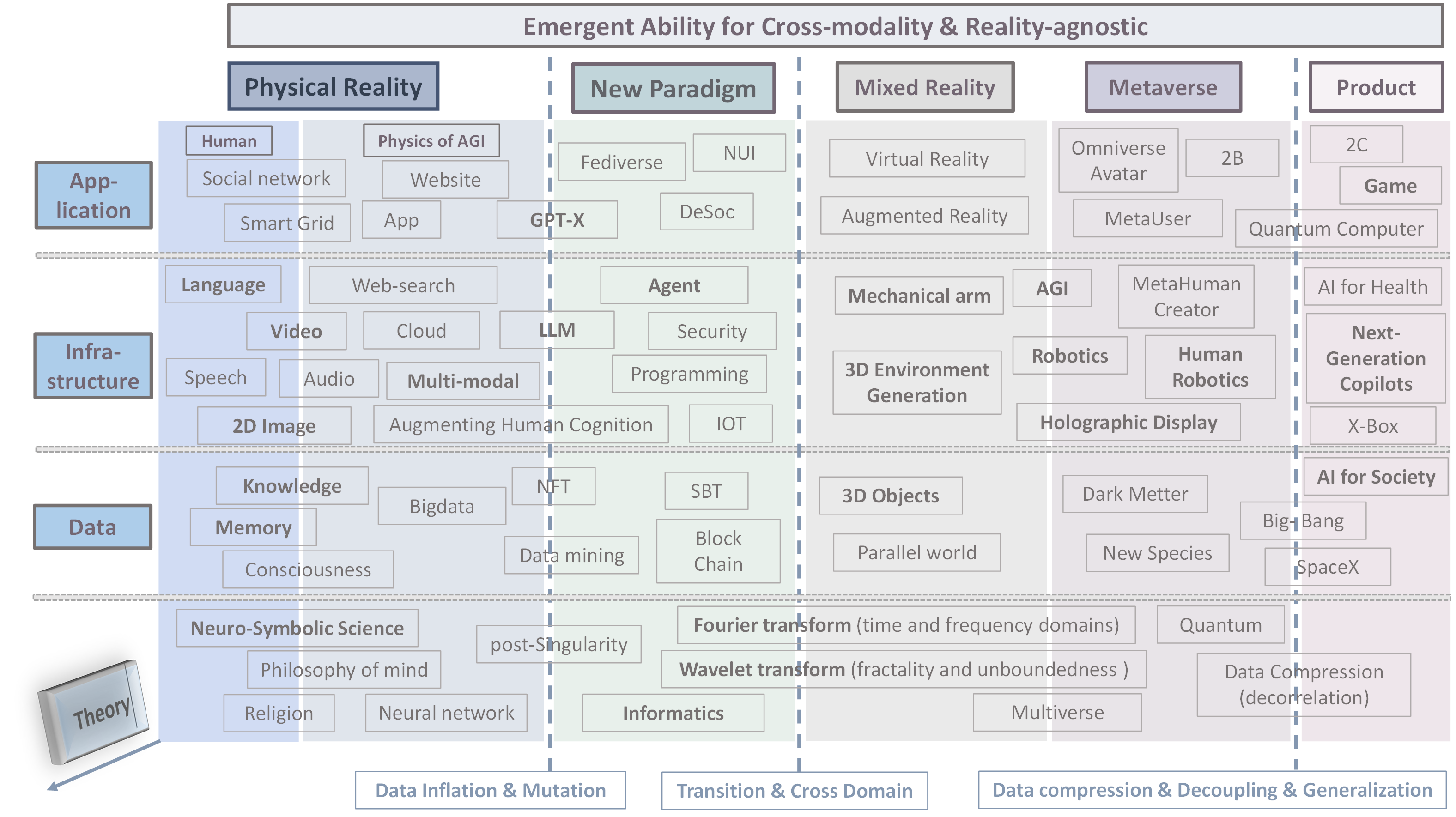}
    \captionof{figure}{The observations about emergent ability. The pipeline shows the emerging capabilities of large foundation models for the cross-modality in a requested unseen environment, and loaded the accountability in reality-agnostic scenario automatically with a new paradigm. 
    We present an AI dominating demonstration of a system that enables interactive generation and editing of a Gaming/AR environment using a knowledge-enhanced style projection. 
    }
    \label{fig:overview}
\end{center}
}]

\blfootnote{$^*$Equal Contribution. Work done while Jae Sung and Abhinav were interning at Microsoft Research.}
\begin{abstract}
Despite the growing adoption of mixed reality and interactive AI agents, it remains challenging for these systems to generate high-quality 2D/3D scenes in unseen environments. 
The common practice requires deploying an AI agent to collect large amounts of data for model training for every new task. This process is costly, or even impossible, for many domains. In this study, we develop an infinite agent that learns to transfer knowledge-memory from general foundation models (e.g., GPT4, DALLE) to novel domains or scenarios for scene understanding and generation in the physical or virtual world. 
The heart of our approach is an {emerging mechanism}, dubbed \textbf{\emph{Augmented Reality with Knowledge Inference Interaction (ArK)}}, which leverages knowledge-memory to generate scenes in unseen physical world and virtual reality environments. 
The knowledge interactive emergent ability (Figure~\ref{fig:overview}) is demonstrated as the observation learns \emph{i) micro-action of \textbf{cross-modality}}: in multi-modality models to collect a large amount of relevant knowledge-memory data for each interaction task (e.g., unseen scene understanding) from the physical reality; 
and \emph{ii) macro-behavior of \textbf{reality-agnostic}}: in 
mix-reality environments to improve interactions that tailor to different characterized roles, target variables, collaborative information, and so on.  
We validate the effectiveness of ArK on the scene generation and editing tasks. We show that our ArK approach, combined with large foundation models, significantly improves the quality of generated 2D/3D scenes, compared to baselines, demonstrating the potential benefit of incorporating ArK in generative AI for applications such as metaverse and gaming simulation. Emergent ability works invisible. 

\end{abstract}

\section{Introduction}
\label{sec:Introduction}

There has been a growing amount of work on using large language models (LLMs) and large multi-modality models (LMMs) to generate high-quality videos and images based on textual inputs \citep{saharia2022photorealistic,yu2022scaling}. Despite impressive results reported, it remains challenging for users (creators) to fully control the generation process and interactively edit the generated results if they don't meet users' intent.
We envision a future AI system where a creator can interactively create a virtual reality scene that consists of objects that do or don't exist in real world, with the system responding faithfully by leveraging knowledge learned from training data of real-world tasks.
For example, an interactive AI agent can incorporate contextual memory and background information, pertaining to a task, into the system by transferring knowledge encoded in pre-trained LLMs/LMMs and multi-sense information collected by sensors during the cause of preforming the task. 
LMMs and LLMs (foundation models) like DALLE-2 and ChatGPT have a superior capability to solve multimodality and natural language reasoning tasks. But we are not yet able to deploy them in many mission-critical real-world applications (e.g., Bing-search, business analysts, office users). Specifically, existing LLMs cannot always effectively transfer knowledge learned from training data to new mission-critical tasks~\citep{peng2023check}. Nor can these models easily
solve complex real-world tasks that require reasoning where human and AI agents often need to collaboratively break a complex task into simpler sub-tasks. 

\begin{figure}[tb]
\centering
\includegraphics[width=0.50\textwidth]{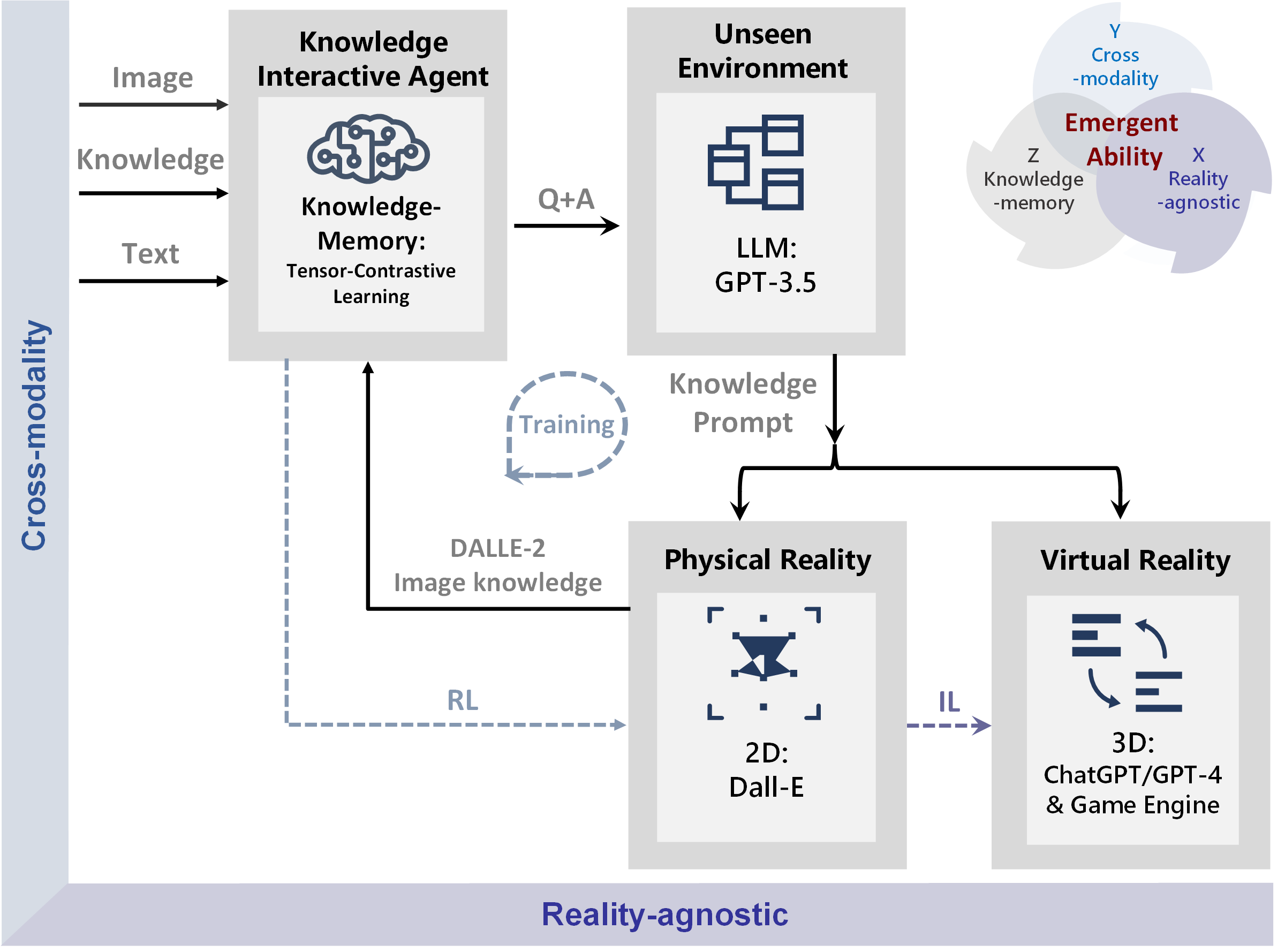}
\caption{Example of ArK Interactive Emergence Mechanism using an external knowledge agent to identify text relevant to the image from a set of candidates.
Our task involves leveraging visual and text knowledge retrieved from web and human-annotated knowledge samples to incorporate external knowledge about the world.}
\label{fig:ArK-flow}
\vspace{-0.1cm}
\end{figure}


\begin{figure*}[htb]
\centering
\includegraphics[width=0.98\textwidth]{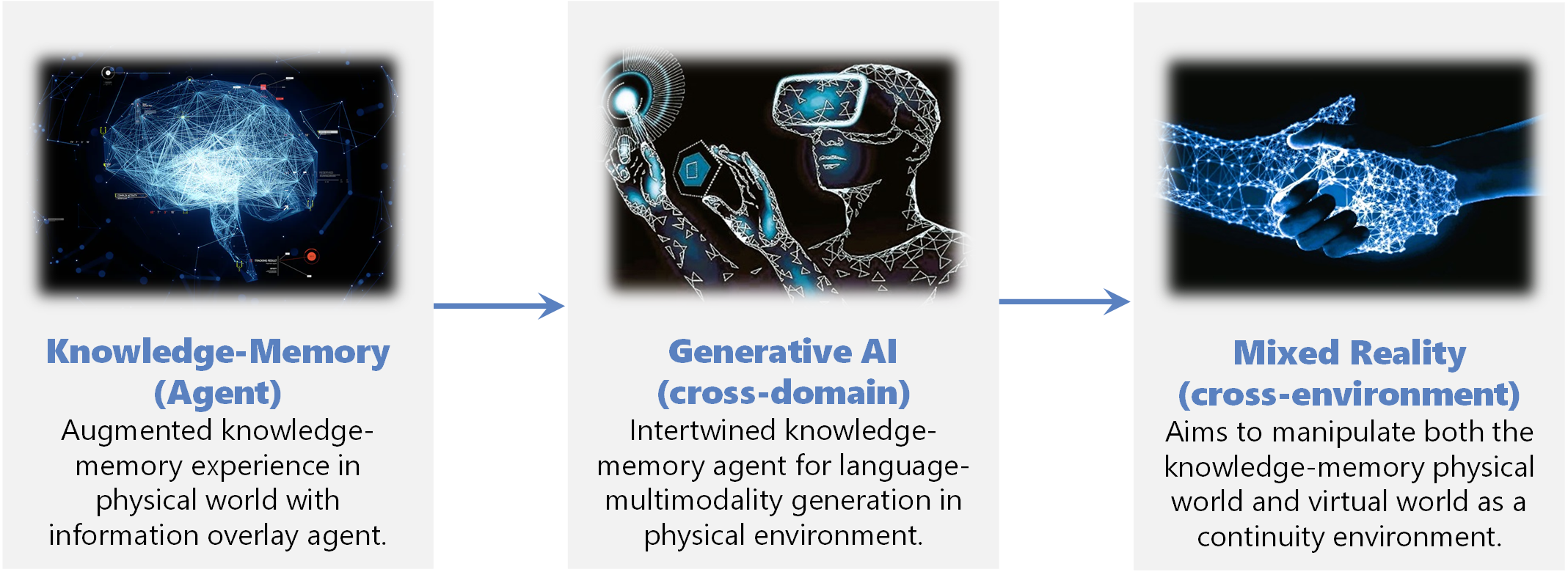}
\caption{Overview of the generative AI with the knowledge memory agent in the physical world and virtual environment.
}
\label{fig: AR}
\vspace{-0.15cm}
\end{figure*}

This study focuses on developing an interactive AI agent for scene understanding and generation, powered by pre-trained foundation models (e.g., DALLE-2, Chat-GPT). It is crucial for the AI agent to not only generate static scenes, but also predict the behaviors of various objects in the generated scene. To this end, the agent needs to retrieve and transfer the knowledge stored in the foundation model to the setting where the scene is being generated, interactively collect external multi-sense information (provided by human creators), and most importantly perform reasoning to synthesize the above two to generate or understand a scene. The reasoning capability of the agent is learned from relevant examples on-the-fly (in-context learning). Due to the length limit of input, we resort to a retriever to retrieve such examples on the fly via e.g., calling the APIs of the external knowledge bases that store such examples. We also need to access the repository of 2D/3D models which can generate 2D/3D objects in the scene. In addition, other knowledge bases, which store meta-information, descriptions, and use cases of these 2D/3D objects, are also useful.

To facilitate human-AI interaction, we have developed a \emph{knowledge-memory} agent, which uses an emerging mechanism, dubbed \emph{Augmented Reality with Knowledge Inference Interaction (ArK)}, for generating and understanding scenes in virtual or real worlds. Specifically, for any particular scene generation or understanding task, the related world \emph{knowledge} is retrieved from a pre-trained foundation model and transferred to the scene, and the \emph{memory} module stores human-AI interactions from which the user intent, or the spec of the scene, can be decoded. Thus, the scene is generated or understood by reasoning over the knowledge and memory.


To demonstrate the effectiveness of \acro{}, we validate our AI agent on four interactive scene understanding and generation tasks: conversational 2D-image generation in physical world, conversational 3D-scene creating in virtual environment, conversational 3D-scene editing in mixed reality, and interactive gaming simulation scenario. Experiments show that \acro{} is effective in collecting and synthesizing knowledge and memory for scene understanding and generation in different settings, according to human evaluation.

Our contributions can be summarized below. 
i) We have developed an infinite knowledge-memory agent for scene understanding and generation in the physical world, and which learning knowledge properties and inference relations that are hard to recover from unseen datasets in virtual reality environments;
ii) We show that the effectiveness of our infinite agent is attributed to the proposed ArK mechanism with reinforce learning (RL) that can understand and generate scenes in unseen settings by effectively synthesizing world knowledge encoded in foundation models, external knowledge retrieved from knowledge bases (e.g. wiki, Conceptnet), and contextual memory collected via human-AI interactions; 
iii) We simulate 3D virtual scenes with imitation learning (IL) in gaming/VR scenario from the 2D knowledgeable purpose applications (2D->3D) in cross-reality. We present experiments and analysis to demonstrate the effectiveness of our approach; and iv) We observed that the explosion of the overall model works efficiently in the cross-modality and agnostic-reality, which depends on the effect of emergent ability in large foundation models. It enhances the interpretation of the existing deep learning model, optimizes the limitations of the unseen environments and unifies the abundant knowledge-memory projection in generative AI system. We make the project and models publicly available \footnote{\url{https://augmented-reality-knowledge.github.io/}}.

\section{Related Work}
\label{sec:RelatedWork}
Recently, text-prompted image generation models such as DALLE-2/GPT4, have been shown to generate images with higher fidelity and relevance to the text query than the image search results and can provide more personalized components (image editing, style change, variations of original image). While these systems can impressively map precise descriptions directly to image, they often fall short of understanding the creator’s intention implied in description in an underspecified text query.

\paragraph{Emerging mechanism in LLM.} Emergent abilities in large language models is one of its characteristic feature that cannot be predicted simply by extrapolating the performance of smaller models. There are several different types of emergent abilities that have been observed in LLMs. One type of emergent ability is the ability to generate creative text formats. For example, LLMs have been able to generate poems, code, and email \citep{wei2022emergent}. Another type of emergent ability is the ability to translate languages. LLMs have been able to translate text from one language to another with a high degree of accuracy. Finally, LLMs have also been able to perform different kinds of tasks, such as answering questions in a comprehensive and informative way.

The exact mechanisms by which LLMs develop these emergent abilities are not fully understood \citep{saharia2022photorealistic,yu2022scaling}. However, it is thought that when LLMs are trained on a large corpus of text, they are able to learn the patterns that exist in language. This allows them to generate text that is similar to human-generated text, and to translate languages with a high degree of accuracy.

\paragraph{Vision-Language transformer.} Multi-modal representation learning is essential for joint vision-language tasks, such as image captioning, visual question answering, and visual commonsense reasoning. Large-scale architectures based on Transformers \citep{transformer} have achieved impressive performance by pretraining representations for a wide range of natural language processing (NLP) tasks \citep{dcwr, devlin2018bert, xlnet, roberta, gpt2}. Recent work on vision-language pretraining (VLP) has shown that these large-scale pretraining methods can also be used for effective cross-modal representations \citep{vibert, lxmert, vlp, uniter, fusion-of-vqa,li2019unicoder, li2019visualbert, oscar, oscarplus, vilt}. Most methods have two stages. First, the architecture is pretrained using a large set of image-text pairs. Then the model is finetuned on task-specific vision-language tasks. For example, \citet{vibert,lxmert} propose multi-stream Transformer-based frameworks with co-attention to fuse these modalities. \citet{vlp, uniter, fusion-of-vqa,li2019unicoder, li2019visualbert, oscar, oscarplus} propose unified pretrained architectures to work on both visual-language understanding and visual-language generation tasks. \citet{conceptbert} uses ConceptNet knowledge graph as is a knowledge base in order to facilitate commonsense vision-language question-answering. \citet{vilt} introduces a pretraining approach to learn self-attention representations directly on image patches. Although these models achieve impressive results on standard vision-language tasks, they do not use information from external knowledge graphs. Our proposed \acro{} architecture shows how the knowledge and reasoning information extracted from text and image facilitates learning more robust and knowledge-aware representations for vision-language tasks.

\begin{figure*}[tb]
\centering
\includegraphics[width=0.9\textwidth]{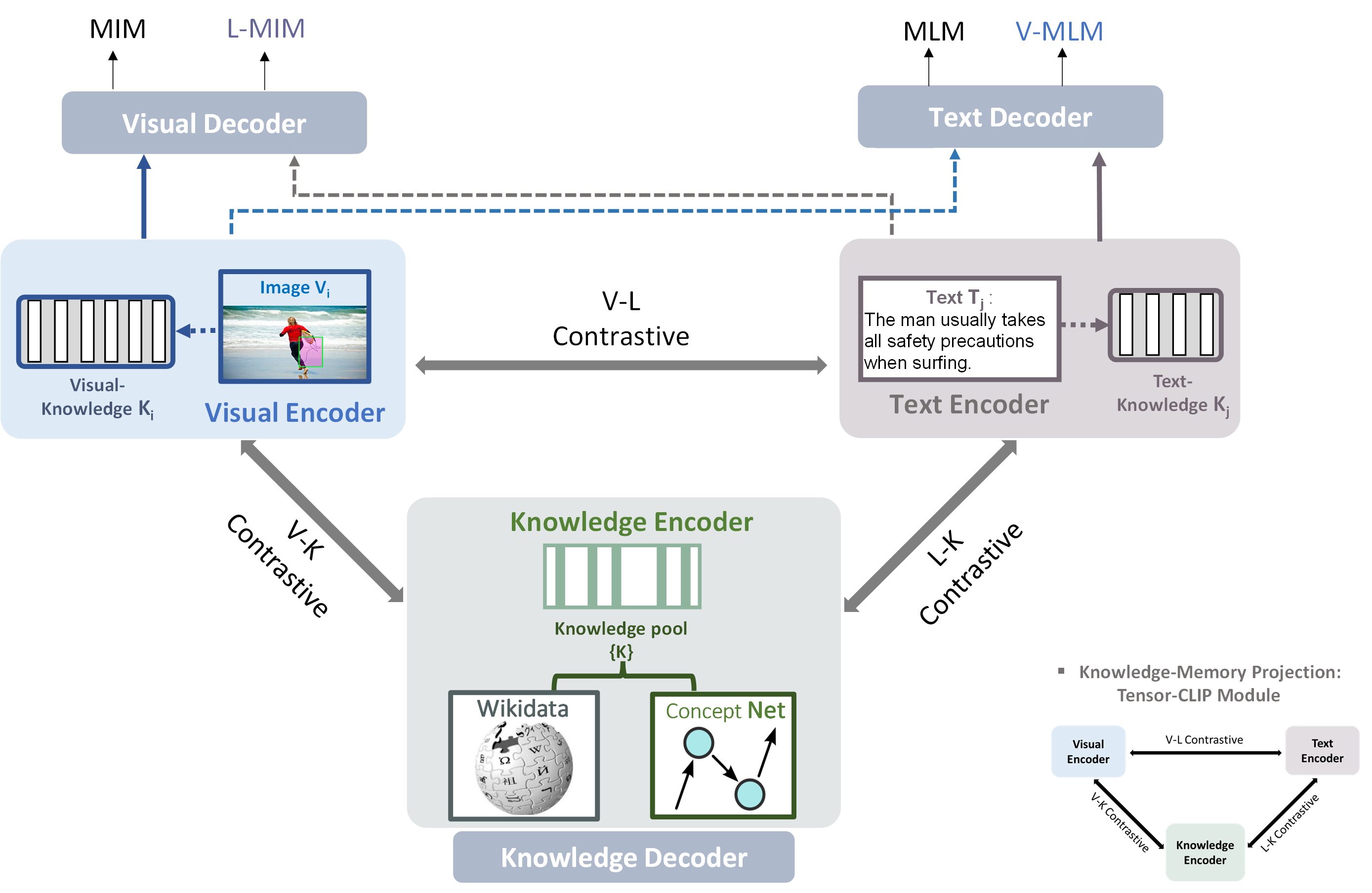}
\caption{Overview of training \textit{Knowledge-Memory Projection}: Knowledge-Tensor-CLIP module. We use the tensor-CLIP to do the linking from the image patches and text phrases to the wikidata and concept entity. We
acquire the positive knowledge for the masked image and text tokens with the nearest neighbor search, and apply weighted contrastive objective. The image and text encoders additionally go through decoder model trained with masked modeling losses respectively. Please find the Section~\ref{sec:app-tensor-clip} for knowledge-memory pretraining, and refer the  Section~\ref{sec:knowledge-memory} for more details the knowledge-agent finetune.}
\label{fig:knowledge-Memory Agent}
\end{figure*}

\paragraph{Language transformer models with knowledge inference.} 
Numerous papers have injected knowledge into language pretraining models \citep{jaket,kplug,corby2021,Zhou2020,kgplm,pencyclopedia,BERT-MK,google-kgml} with an emphasis on NLP tasks. For example, \citet{jaket} extracts knowledge graph information from Wikipedia, and uses it to help the pretraining progress. \citet{kplug} injects domain-specific knowledge in pertraining language models for NLP tasks. These methods focus on language tasks, and have not been applied to multi-modal transformers (e.g. for vision and language). More recently, KRISP (\citet{KRISP}) was proposed to retrieve implicit knowledge stored in pre-trained language models as a supplementary knowledge resource to the structured knowledge base. MAVEx (\citet{MAVEx}) presented an answer validation approach to make better use of the noisy retrieved knowledge. Additionally, some proposed structures and representations are domain-specific and are hard to extend to new tasks. In this paper, we introduce a knowledge-based pretraining model that uses the transformer architecture for multi-modal understanding and reasoning. The knowledge representations in our method can be easily extracted from massive data. 

\begin{figure*}[htb]
\centering
\includegraphics[width=0.99\textwidth]{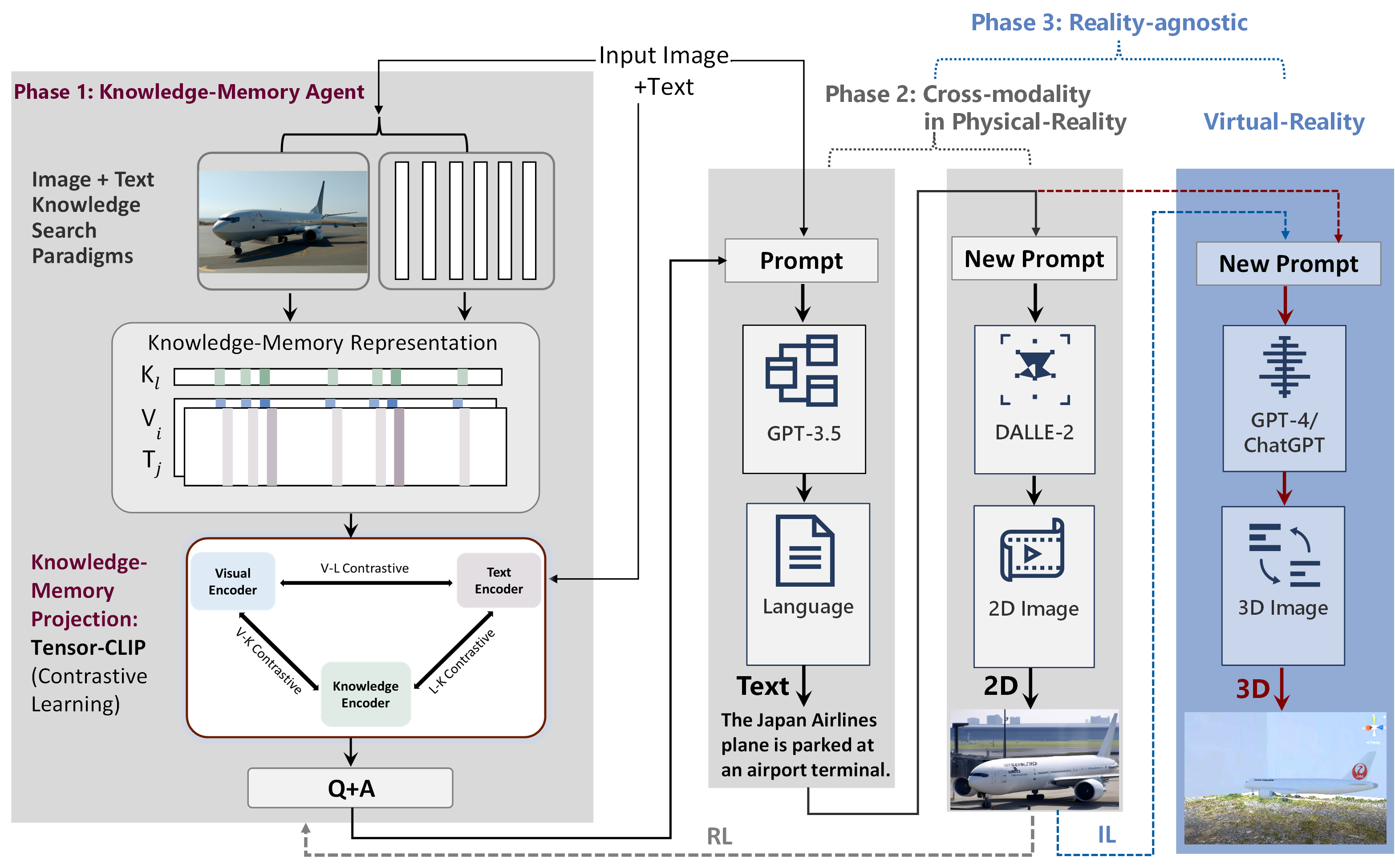}
\caption{The \acro{} model: At training time, the agent retrieves relevant knowledge for the given image-text pair and asks question and answer for the query. The question and answer are provided to large language models (GPT-3.5, ChatGPT) that generates a new prompt that would be called by the DALLE-2 model. The similarity between the generated image and original image used as reward to train the agent to learn to select relevant knowledge, while the blackbox models are kept as frozen. At test time, we generate the 2D images with the input text, and model follows the same loop until the new prompt generation step. Instead of feeding back to DALLE-2 for the new prompt, we use ChatGPT to generate a code snippet runnable in a 3D rendering enginne such as Unity. Overall, we use the external knowledge and the visual priors from the generated 2D image to improve the 3D scene generation.
}
\label{fig:ArK}
\vspace{-0.3cm}
\end{figure*}







\section{Knowledge Source}

\begin{figure}[tb]
\centering
\includegraphics[width=0.49\textwidth]{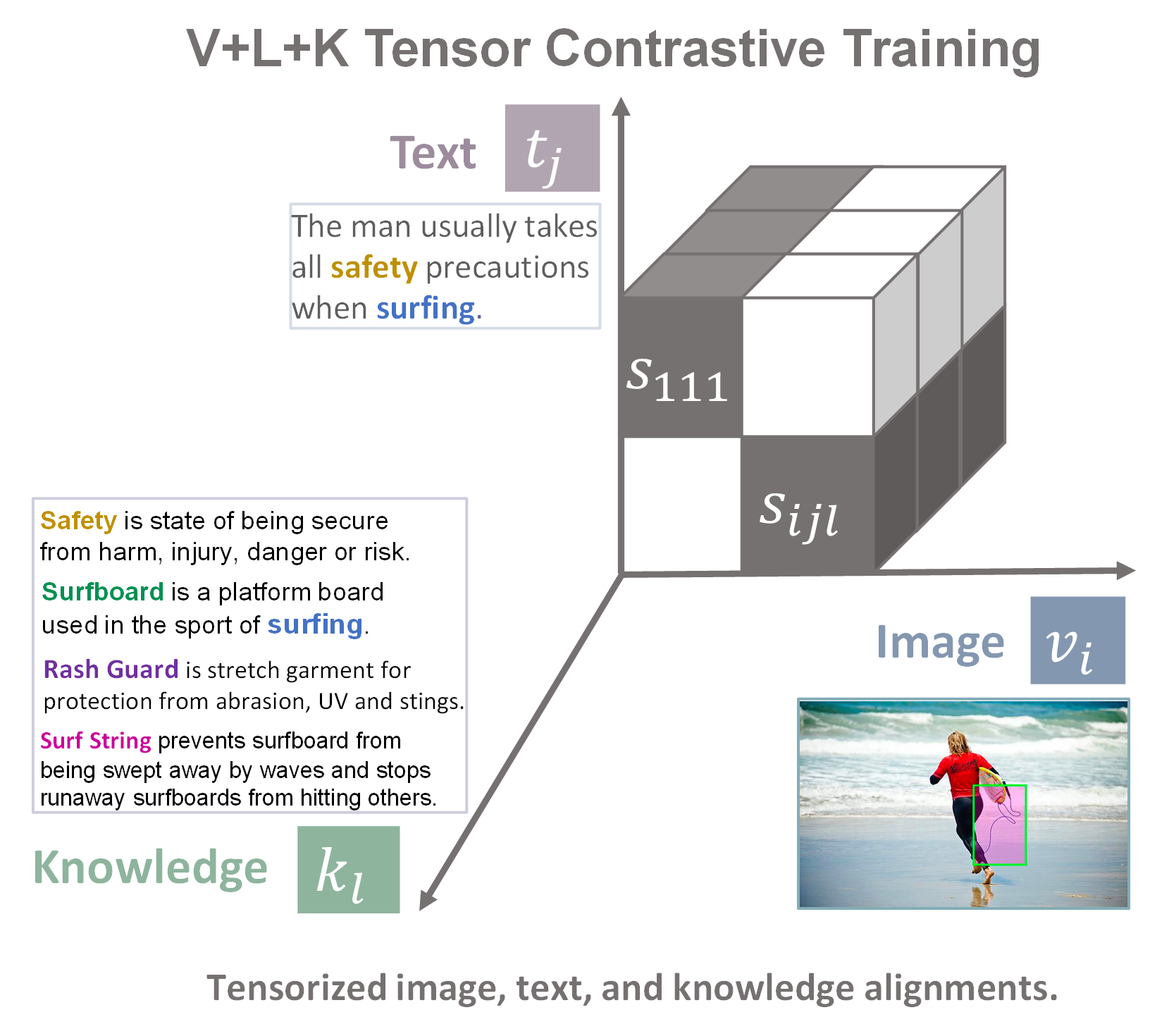}
\caption{Example of ArK task that uses knowledge to identify the relevant knowledge to the image-text candidates.
Our task involves leveraging visual and text knowledge retrieved from web-search and implicit knowledge from Open-AI foundation models and the incorporate external web-search knowledge pool about the world.}
\label{fig:ArK-knowledge}
\vspace{-0.1cm}
\end{figure}

\subsection{Implicit Knowledge Source from OpenAI Models}
\paragraph{Text retrieval knowledge: large language models (GPT-3.5).}

\citet{yang2021empirical} propose to use GPT-3 for the outside knowledge-based visual question answering task OK-VQA.
Instead of using explicit knowledge sources, they use GPT-3 as an implicit source of knowledge.
They propose to feed the question and textual descriptions of the image to the GPT-3 model and query it to directly predict the answer.
Their model improved the state-of-the-art on OK-VQA by a significant margin of over $9\%$.
Their qualitative analysis shows that the KAT model works quite well on various questions that require external knowledge. Thereby demonstrating the implicit knowledge contained in GPT-3.

Following the KAT ~\cite{gui2022kat} and PICa model, for each image-question pair, we construct a carefully designed text prompt consisting of a general instruction sentence, the description of the image, the question, and a set of context-question-answer triplets taken from the training dataset that are similar to the current image-question pair. We then input this text prompt to the GPT-3.5 model in its frozen version and obtain the output from GPT-3.5 as the tentative answer candidate to the current image-question pair.


\paragraph{Image retrieval knowledge: image generation module (DALLE-2).}
DALLE-2 \cite{ramesh2022hierarchical} and Stable Diffusion \cite{rombach2021highresolution} are text to image generation models that can fill in the scene context via visual inductive bias, which can be considered as source of implicit visual and physical knowledge. For example, one can visualize how the chairs are orientated when there are four chairs around the room, or what objects are typically present in a video conference room. We leverage this model as our knowledge source by running vision language model to extract information from generated image. Specifically, for the task of 3D scene generation, we first generate the 2D image that contains informative scene prior and use the generated 2D content (e.g. orientation, environment objects) to help guide the 3D scene generation.

\subsection{Explicit Knowledge Source from Web Knowledge Bases}
\label{section: knowledge_source}
We describe the explicit knowledge source to train the Knowledge Tensor-CLIP module. This pool is additionally used for the knowledge-memory agent to retrieve the relevant knowledge for the image and text pairs, and generate new, enhanced prompt for cross-domain space generation.

\paragraph{Factuality knowledge: Wikidata.}
Wikidata \cite{wikidata} is an open web-based knowledge base of real-world entities. We use the cleaned version of entity and description text, and format the knowledge text as ``\texttt{\{entity\} is a \{description\}}'' following \cite{wang2021kepler}\footnote{\url{https://deepgraphlearning.github.io/project/wikidata5m}} and further filter non-English entities, resulting 3,836,524 sentences in total.  

\paragraph{Commonsense knowledge: ConceptNet.}
ConceptNet \citep{Conceptnet} is a crowd-sourced project with over 34 million facts organized as knowledge triples collected by translating English language facts into an organized triple structure. It inherently supports common sense knowledge of semantic concepts such as (dog, has property, friendly). We use the dump in Conceptnet 5.5~\footnote{\url{https://huggingface.co/datasets/conceptnet5}}, and extract 7 types of relation knowledge for English word concepts (IsCapableOf, HasProperty, Causes, AtLocation, PartOf, MadeOf, UsedFor).  In total, we obtain 2,697,499 unique triples.

\section{Approach}

\paragraph{Augmented physical and virtual scenes generation with knowledge-memory agent.} The framework of interactive text to 3D scene generation is shown in Fig. \ref{fig:ArK}, which extends the paradigm of calling blackbox models with a trained agent that actively seeks to collect knowledge useful for scene generation. Here, the blackbox models are not trained, and we improve their performance by providing improved text prompts at test time. This involves a knowledge-interactive modeling through a combination of triple systems - one performing knowledge retrieval from image and text query, second performing question and answer generation from the relevant knowledge, and last one writing a new, informative prompt with reinforcement learning. At test time, we then generate the 2D image output by the three systems and run a final RL process with another knowledge-enhanced DALLE-2/ChatGPT query model to obtain our final image to an image-question pair. Below we describe the details of our proposed triple systems and how we combine the outputs to generate the final 3D image through finetuned knowledge-agent and zero-shot models.

The whole model is trained with three phases.
1) Knowledge-memory agent module for self-supervised Learning; 2) Reinforcement learning module for 2D sense generation in physical world; 3) Imitation Learning module for virtual environment generation.


\subsection{Phase 1: Knowledge-memory Agent Training with Self-supervised Learning}
\label{sec:knowledge-memory}
Next, we will introduce our trained Knowledge-Memory Agent as the first phase.

\paragraph{Knowledge retrieval system.}The knowledge retrieval model takes in the image $I$ and the text caption $T$ to retrieve the useful knowledge statement $k^*$ that aids the understanding of both image and text. 
This retrieved knowledge statement is used as additional context for powerful instruction finetuned language models such as GPT-3.5 to rewrite the text query appropriately.

\paragraph{Training Knowledge-Tensor-CLIP module.}

For the knowledge retrieval system, as shown in Figure~\ref{fig:ArK-knowledge}, 
we introduce Knowledge-memory tensor CLIP, a novel image-text-knowledge module that leverages explicit knowledge as bridge to connect the vision and language modalities. The vision encoder is initialized with the CLIP ViT-B/16~\cite{radford2021learning} visual encoder model, and the text and knowledge encoder are initialized with the text encoder model.

To extract knowledge, we follow KAT~\cite{gui2022kat}, in which the image and text pairs are represented as dense vectors, computed by the image and text encoder of frozen CLIP model. A single maximum inner product search (MIPS) index is then built using FAISS~\cite{johnson2019billion} to perform nearest-neighbor search. In our setup, we have three dimensions of embeddings (image $V$, text $T$, and knowledge $K$). During training, we keep the knowledge encoder as frozen, while the image and text encoder are updated due to the computational cost to update the knowledge index at every step of training\footnote{Because the image and text encoders in CLIP have been pre-trained with contrastive objective, running MIPS at initial training will still retrieve relevant knowledge.}. Thus, we create fixed index embeddings using the frozen knowledge encoder model.

To train the model, we use the contrastive losses used in ~\citet{radford2021learning}. As shown in Figure~\ref{fig:knowledge-Memory Agent}, the model is trained with three-way contrastive learning objectives: (Vision-Language, Language-Knowledge, and Image-Knowledge). The vision-language direction loss $L_{v2t}$ and $L_{t2v}$ follows the original objective in CLIP. To acquire the positive knowledge for image-knowledge and language-knowledge direction for the $i$th batch, we retrieve the top-k knowledge from image $K_{v^i}$ and text $K_{t^i}$ respectively with nearest neighbor search. The $k$ retrieved knowledge are given the positive label, and the ones from different batch are labeled as negative. If we define $u$ as the knowledge vector, $v$ as the visual vector, and $w$ as the text vector, the model is trained with the loss $L$ with the weighted ($a,b,c,d$) contrastive losses:

\begin{align}\label{eq:knowledge-image loss}
{L}_{v2k} &=  \sum_{i \in B}{\log \frac{
\sum_{k \in K_{v^i}}{u^T_k v_i}} {\sum_{k \in \{K_{v^1}, ... K_{v^i}, K_{v^B}\}}{u^T_k v_i}}}
\end{align}

\begin{align}\label{eq:knowledge-text loss}
{L}_{t2k} &=  \sum_{i \in B}{\log \frac{
\sum_{k \in K_{t^i}}{u^T_k w_i}} {\sum_{k \in \{K_{t^1}, ... K_{t^i}, K_{t^B}\}}{u^T_k w_i}}}
\end{align}

\begin{align}\label{eq:knowledge loss}
L_{cont} &=  a{L}_{v2t} + b{L}_{t2v} + c{L}_{v2k} + d{L}_{t2k}
\end{align}

Following ~\cite{beit3}, we further apply masked image ($L_{MIM}$), language ($L_{MLM}$), and vision-language ($L_{MVLM}$) modeling losses additionally to the image and text encoder based on their effectiveness in the pre-training stages. BEIT-2 is used to get the masked image labels.

\begin{align}\label{eq:knowledge loss}
L_{mask} &=  L_{MIM} + L_{MLM} + L_{MVLM}
\end{align}

The final loss to train the Knowledge-Tensor CLIP module is $L = L_{cont} + L_{mask} $. We refer to Section~\ref{sec:app-tensor-clip} in the appendix for more pre-training details. The full training framework is shown in Figure~\ref{fig:knowledge-Memory Agent}.

\paragraph{Inference.}
 At test time, along with the image, we consider extracting knowledge for the individual noun phrases rather than for the entire sentence. This is to ensure that different knowledge for the mentioned objects is extracted that are seldom ignored if only the global sentence context is considered to extract knowledge. To do so, we extract $p$ noun phrases $W_{0,...p}$ with parser tools such as Spacy, and acquire $p$ phrase embeddings $e_{0,...p}$. We then acquire the visual embedding $v$, and use the average of phrase and visual embeddings from CLIP: $\alpha e_i + (1-\alpha) v$ as query $q_i$ to perform the nearest neighbor search. We set $\alpha$ as 0.5 and we pick the top-1 best phrase knowledge as our external knowledge based on the cosine similarity score. 
 We evaluate the Knowledage-Tensor-CLIP model on different dataset, and show the result in Section 5.1.  


\subsection{Phase 2: Knowledge Enhanced Physical Scene Generation with RL}
Next, we wish to incorporate the knowledge source to generate a new prompt that contains informative content for physical QA-2D scene generation. After the agent retrieves the relevant knowledge for the given image, text pair, it generates a question and answer tuple using the retrieved knowledge. This model is trained using Reinforcement Learning and is described in the following sections.

\paragraph{Learning knowledge-memory agent.} In the first phase of supervised training, we first train the model to ask questions and answer on visual question answering dataset, such as AOKVQA.

Since the parameters of the LLMs such as GPT-3.5 are frozen, during training, the agent receives no information to learn if the retrieved knowledge and generated QAs are indeed useful for the downstream task. Hence we use the feedback from generated images with the knowledge prompt to train the agent using reinforcement learning. We use policy gradient algorithm~\cite{Sutton1999PolicyGM} to train the agent with the reward from similarity between original image and image generated with knowledge enhanced prompt (Red direction in Figure~\ref{fig:ArK}. The image is generated with DALLE and leverages the image-knowledge source to train the agent.

\paragraph{Physical scene generation with knowledge enhanced prompt scheme with RL.} 

After the agent retrieves the relevant knowledge using the knowledge retrieval model $K(V,T)$ for the image $V$ and text $T$, it generates a question and answer using the retrieved knowledge and image. We use the knowledge-based visual question answer dataset, AOKVQA, as supervision text and image retrieved knowledge to apply reinforcement learning. 
We further augment the question and answer pairs by prompting GPT-3.5 to generate question and answer using the $k$ retrieved knowledge. The prompt is given as: \texttt{Original Sentence: \{\} Knowledge: \{\}. Generate question and answer relevant to the sentence and knowledge.} (The details please find in the Figure~\ref{fig:prompt1} for prompt in Appendix). This way, the augmented question-answer pairs has the size of $k$ times the size of AOKVQA data, and we use $k = 5$ in our experiments.
With this supervision, we then train the agent with seq2seq objective that asks the relevant question and answer in convectional way for both 2D image and knowledge text in knowledge-memory reinforcement learning. Next, we use GPT-3.5 to reformulate the text query using the new knowledge and question-answer with the following prompt: $\texttt{Original Sentence: \{\} Question: \{\}}$ $\texttt{Answer: \{\}  New Sentence:}$. (see Figure~\ref{fig:prompt2} for prompt template).
This Phrase to prepare for the virtual 3D scenes generation, which with the 2D retrieved image from DALLE, new prompt text and w/knowledge from GPT-3.5, and our convectional question-answer pairs from our trained knowledge memory agent.


\paragraph{Reinforcement learning using feedback.}
In the first stage of the training, the agent gets no signal from the blackbox model such as ChatGPT and GPT-3.5 to know if the retrieved knowledge and generated QAs are indeed useful for the blackbox models as their weight are frozen. 
Based on the previous application of reinforcement learning to Natural Language Generation models, we consider the agent to be a policy $\pi_\theta$ with generated question answer sequence $qa_k$ as state. To train the model with reinforcement learning, we use the feedback from generated images with the knowledge prompt $k$ to train the agent. Specifically, we use policy gradient algorithm~\cite{Sutton1999PolicyGM} to train the agent with the reward $R$ from the cosine similarity between original image $V$ and image generated with knowledge enhanced prompt $\Tilde{V_k}$ measured by the CLIP ViT-B16 visual encoder model (Red direction in Figure~\ref{fig:ArK}. Since we cannot compute the partial reward at each generated token or state, the reward is calculated after the sequence has been fully generated. In the end, the reward $R$ for the image $V$ and text $T$ is computed as follows. 
\begin{align}\label{eq:Reward}
qa_k &= \pi_\theta(\text{K}(V,T), V) \\
\Tilde{T_k} &= \text{GPT-3.5}(T, qa_k) \\
\Tilde{V_k} &= \text{DALLE-2}(T_k) \\
R(V, T) &= \text{cos}(CLIP(V), CLIP(\Tilde{V_k}))  
\end{align}

Subsequently, the agent is trained using reinforcement learning to incorporate the feedback using the reward. We use the actor-critic algorithm PPO~\cite{schulman2017proximal} to update the parameters of the agent using its clipped version:

$$\hspace{-50mm}L_{\text{CLIP}}(\theta) =$$
\begin{align}\label{eq:parameters}
\hspace{-5mm}\mathbb{E}_t\left[\min\left(R_t(\theta)\hat{A}_t, \text{clip}(R_t(\theta), 1-\epsilon, 1+\epsilon)\hat{A}_t\right)\right]
\end{align}
Here $\epsilon$ is a constant which is set to 0.2 and $\hat{A}$ refers to the advantage estimate.



\subsection{Phase 3: 3D Virtual Scenario Diagram with Imitation Learning}
In the third phase of the training, the trained knowledge agent is used to perform 3D scene generation. Note the agent requires an image and original text to generate relevant knowledge for the query. Since the image is not provided at test time during scene generation, we use text-to-image generation model DALLE-2 to reconstruct the 2D anchor image which is further used to extract the desired knowledge. Here, DALLE-2 implicitly serves as the image-knowledge source that contains the visual prior knowledge of what we can imagine from the text query. The agent then takes as input the original text and the generated 2D image to retrieve knowledge and outputs a question and answer tuple (e.g., Figure~\ref{fig:3d_example}), while GPT-3.5 generates new knowledge-enhanced prompt using the agent output. 

To generate the 3D scene from knowledge prompt, we use GPT-4/ ChatGPT to output text code that is then rendered using a 3D rendering engine. We use the prompt and code syntax in GPT-4/ ChatGPT to generate the spatial arrangement in the Unity game engine. We perform experiments with GPT-4/ ChatGPT as the code generation model, and use the Sketchfab API to load the 3D models viewable in the Unity game engine. More information about generating the prompt to run the Unity game engine can be referenced in \citet{Roberts2022StepsTP}.
\begin{equation}
\theta^* = \arg\max_{\theta} \frac{1}{N} \sum_{i=1}^N \log p_{\theta}(a_i|s_i)
\end{equation}
where $\theta$ refers to the parameters of the model, $N$ is the number of demonstration trajectories, $s_i$ and $a_i$ are the state and action at time step $i$. The objective of imitation learning is to find the optimal policy parameters $\theta^*$ that maximize the log-likelihood of the expert demonstrations.


\subsection{Emergent Ability Survey for Cross-modality and Reality-agnostic Observation/Discussion}
We train the knowledge-memory agent and use an infinite feedback loop with RL in the real world to randomly initialize a policy, but this strategy does not work well with virtual reality where it is difficult to obtain initial rewards in the 3D environments or the unseen environment with 2D ground truth image, especially in the virtual environments with sparse rewards or only terminal rewards. Therefore, a better solution is to use a trained infinite-memory agent with expert characteristics through imitation learning to help the agent explore better and utilize the unseen environmental space. Imitation Learning~(IL), can learn policies directly from expert data. 

\paragraph{IL -> Decoupling.}
In traditional imitation learning, an agent learns a policy by mimicking the behavior of an expert demonstrator. However, directly learning the expert policy may not always be the best approach, as the agent may not generalize well to unseen situations. To address this issue, we propose learning an implicit reward function that captures the essential aspects of the expert's behavior as shown in Phase 2. This results in giving the infinite knowledge-memory agent some information about physical-world behavior to perform tasks by learning from expert demonstrations. It addresses some limitations of traditional imitation learning, which often requires a large amount of expert data and may suffer from compounding errors in complex tasks. 


\begin {table*}[t]
\normalsize
\centering
\scalebox{0.6}{
\begin{tabular}{l | c | c | c c | c c | c | c }
\toprule
{\textit{Knowledge Category}} & \multicolumn{2}{c}{\textit{Semantic
}} & \multicolumn{2}{c}{\textit{Encyclopedic}} & \multicolumn{3}{c}{\textit{Commonsense
}} & \multicolumn{1}{c}{\textit{Open-World
}}\\
\cmidrule(lr){2-3}\cmidrule(lr){4-5}\cmidrule(lr){6-8}\cmidrule(lr){9-9}

\textbf{Dataset} & \multicolumn{1}{c}{\textbf{Coco}} & \multicolumn{1}{c}{\textbf{Flickr 30K
}}& \multicolumn{2}{c}{\textbf{WIT}} & \multicolumn{2}{c}{\textbf{Sherlock}} & \multicolumn{1}{c}{\textbf{VisualCOMET}} & \multicolumn{1}{c}{\textbf{AOKVQA}} \\

\midrule
\rowcolor{gray!5} \textbf{Approach} & Text -> Image & Text -> Image & Text -> Image & Text -> Image & Text -> Image & Image -> Text & Text -> Image & Mult. Choice \\
\rowcolor{gray!5} & Zero Shot & Zero Shot & Zero Shot & Zero Shot & Finetuned  & Zero Shot &  Finetuned &  Finetuned \\
\rowcolor{gray!5} \textbf{Metric (\%)} & R@1(\%)$\uparrow$ & R@1(\%)$\uparrow$ & R@1(\%)$\uparrow$ & Rank$\downarrow $ & Rank$\downarrow$ & Rank$\downarrow$ & Acc(\%)$\uparrow$ & Acc(\%)$\uparrow$ \\
\midrule
\rowcolor{gray!10} \multicolumn{9}{l}{\textit{w/o Knowledge}} \\
\rowcolor{gray!10} Contrastive & 49.7 & 51.8 & 42.0 & 21.1 & 28.3 & 28.7 & 53.8 & 60.4 \\

\midrule
\rowcolor{gray!15} \multicolumn{9}{l}{\textit{w/ Knowledge (w/o Mask)}} \\
\rowcolor{gray!15} Knowledge-Tensor-Cont. (ours)
 & 49.8 & 50.9 & \textbf{43.4} & - & 27.4 & - & \textbf{54.5} & 61.2 \\

\bottomrule
\rowcolor{gray!30} \multicolumn{9}{l}{\textit{w/ Knowledge (w/ Mask) \ \ }} \\
\rowcolor{gray!30} \textbf{Knowledge-Tensor-Cont. (ours)} 
& \textbf{50.3} & \textbf{52.0} & \textbf{43.3} & \textbf{15.4} & \textbf{27.2} & \textbf{27.3} & \textbf{54.4} & \textbf{61.3} \\

\bottomrule
\end{tabular}
}
\caption {Results of text \textit{to} image, and image \textit{to} text retrieval of Knowledge-Tensor-CLIP Memory training. We report the average rank of ground truth image/text, Recall@1 (R@1), and Accuracy (acc) measuring if ground truth image/text is retrieved in top $k$ retrieved knowledge.} 
\label{tab:knowledge_retrieval}
\vspace{1.5mm}
\end {table*}

\begin{figure*}[htbp!]
    \centering
    \includegraphics[width=0.98\textwidth]
    {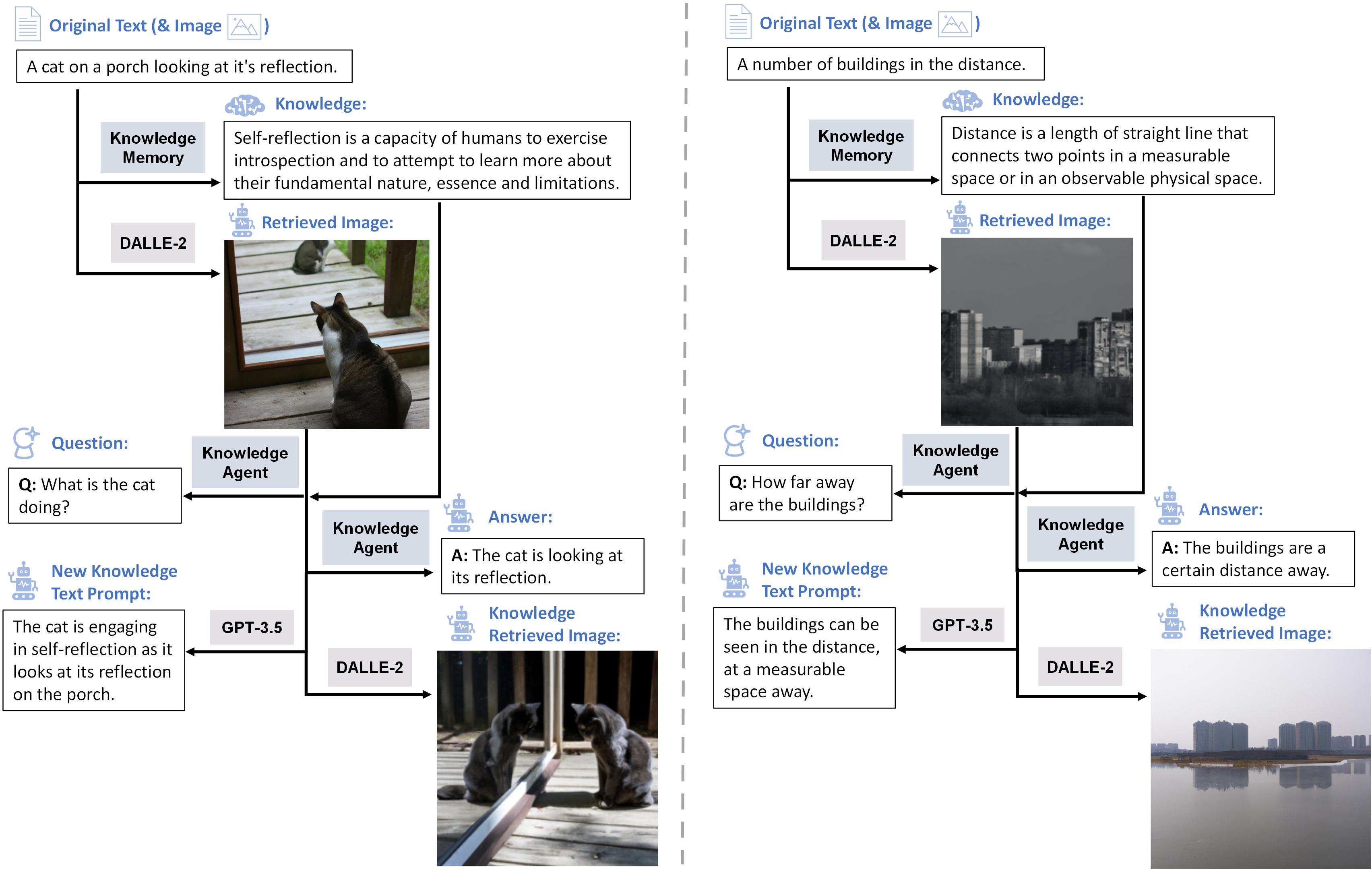} 
    \caption{Qualitative examples of conversational-2D Image generation with Knowledge Enhanced Prompts.}%
    \label{fig:2d_example}%
\end{figure*}

\begin{figure}[tb]
\centering
\includegraphics[width=0.49\textwidth]{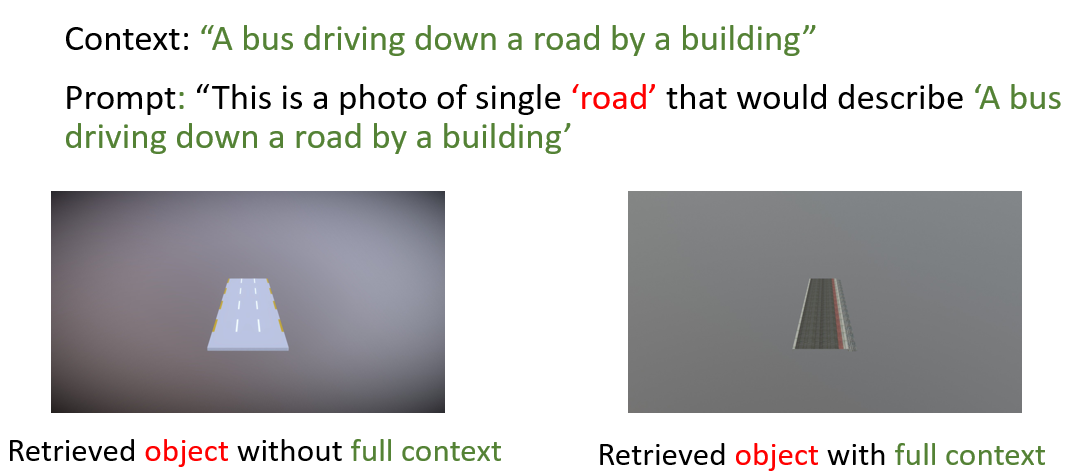}
\caption{Comparison of the retrieved object with and without context. We see that prompting CLIP with the full context retrieves a more appropriate object (asphalt road).
}
\label{fig:objaverse}
\vspace{-0.3cm}
\end{figure}

The key idea behind our IL approach involves two components: 1) the infinite agent in physical-world expert demonstrations are collected in the form of state-action pairs; and 2) the virtual environment imitating agent generator. The imitating agent aims to produce actions that mimic the expert's behavior, while the virtual environment imitating agent learns a policy mapping from states to actions by minimizing a loss function that measures the difference between the expert's actions and the actions generated by the learned policy.


\paragraph{Decoupling -> Achieve Generalization.}
Instead of depending on a task-specific reward function, the agent learns from expert demonstrations. These demonstrations provide a diverse set of state-action pairs that cover various aspects of the task. The agent can then learn a policy that maps states to actions by imitating the expert's behavior. Decoupling in the context of imitation learning refers to separating the learning process from the task-specific reward function. This separation allows the learned policy to generalize across different tasks without explicitly relying on the specific reward function for each task. By decoupling, the agent can learn from expert demonstrations and learn a policy that is adaptable to a variety of situations.

Decoupling enables transfer learning, where a policy learned in one domain can be adapted to other domains with minimal fine-tuning. 
Since the agent does not rely on a specific reward function, it can adapt to changes in the reward function or environment without the need for significant retraining. This makes the learned policy more robust and generalizable across different environments.



\paragraph{Generalization Element -> Emergent Behavior.}

Generalization can be used to explain how emergent properties or behaviors can arise from simpler components or rules. The key idea lies in identifying the basic elements or simple rules that govern the behavior of the system. In the context of artificial intelligence, these elements could be individual neurons, simple heuristics, or basic algorithms. Consequently, by observing how these simple components or rules interact with one another. These interactions often lead to the emergence of more complex, higher-level behaviors or properties that cannot be predicted or explained by merely examining the individual components in isolation. Generalization across different levels of complexity: By learning general principles that apply across different levels of complexity, a system can exhibit emergent properties. This generalization enables the system to adapt and respond to new situations, demonstrating the emergence of more complex behaviors from simpler components or rules. Finally, the ability to generalize across different levels of complexity allows for the transfer of knowledge from one domain to another. This transfer contributes to the emergence of complex behaviors or properties in a new context, as the system adapts.


\section{Experiments and Results}


\begin{table}[t]
\footnotesize

    \centering
    \begin{tabular}{lc}
    \toprule
        {\footnotesize \textbf{Model}} & Text to Image (R@1) \\
        \midrule
        \citet{srinivasan2021wit} & {34.4} \\
        CLIP~\cite{radford2021learning} & 42.0 \\
        Knowledge-CLIP (Ours) & 43.4 \\
    \bottomrule
    \end{tabular}
    \caption{Text to Image Retrieval on WIT dataset trained with WIT-en. Recall@1 (R@1) is reported for the metric. Ref+Attr is as text input following ~\citet{srinivasan2021wit}.}
    \label{tab:wit_dataset}
    \vspace{-0.3cm}
\end{table}

\subsection{Knowledge Agent Training}

\paragraph{Knowledge Tensor-CLIP training implementations.}
We finetune the Knowledge-CLIP model on the WIT dataset~\cite{srinivasan2021wit} with the filtered version to only consider English texts, totaling 5M in training data and 30K on test data. We use Wikidata ~\cite{wikidata} and ConceptNet ~\cite{Conceptnet} as the explicit knowledge source as the bridge to connect the image and text modalities. The Knowledge-CLIP model is trained with batch size of 2048, image size of 224, and the Adam optimizer ~\cite{Kingma2014AdamAM} with $\beta_1 = 0.9$, $\beta_2 = 0.98$, and $\epsilon =1e-6$ for optimization. We use a cosine learning rate decay scheduler with a peak learning rate of 1e-5 and a linear warm-up of 10k steps. The weight decay is 0.05. More pre-training details of the Knowledge-Tensor CLIP model is in Section~\ref{sec:app-tensor-clip} of Appendix.

\paragraph{Knowledge-memory retrieval system and evaluation.}
We first evaluate the performance of knowledge retrieval system on the WIT dataset~\cite{srinivasan2021wit}, Coco~\cite{mscoco}, Flicker 30K ~\cite{plummer2015flickr30k}, Sherlock~\cite{hesselhwang2022abduction}, and AOKVQA~\cite{AOKVQA} on the knowledge-based image text retrieval and question answering task. For WIT training and evaluation, we concatenate the reference and attribute to acquire the text representation following \citet{srinivasan2021wit}\footnote{One of reference or attribute is used if both are not available in the data.}.

In the experiments, we run ablations of the proposed knowledge module and the masking loss in the pre-training stage. We refer to \textit{Contrastive} as the model only on vision-language direction, \emph{i.e.} the same as the CLIP  training objective ~\cite{radford2021learning}, \textit{Knowledge-Contrastive} (ours) as Contrastive with knowledge contrastive loss, and \textit{Knowledge-Contrastive-Mask} (ours) as Knowledge-Contrastive trained with the masking loss. We refer to Knowledge-CLIP as this final model.

 Table~\ref{tab:knowledge_retrieval} and Table~\ref{tab:wit_dataset} present results on image text retrieval on different knowledge categories dataset: Semantic knowledge, Encyclopedic knowledge, Commonsense knowledge, open-world knowledge, comparing the model trained with (Knowledge-contrastive-training) and without knowledge (CLIP). We show analysis plots for loss tendency of training infinite knowledge-memory agent in Appendix~\ref{app:analyzation}.
We see that our knowledge-contrastive-training model provides improvement over the contrastive on the data, for dataset that requires entity-based knowledge. 

\begin{figure*}[htb!]
    \centering
    \subfloat[\centering]{{\includegraphics[width=0.95\textwidth]{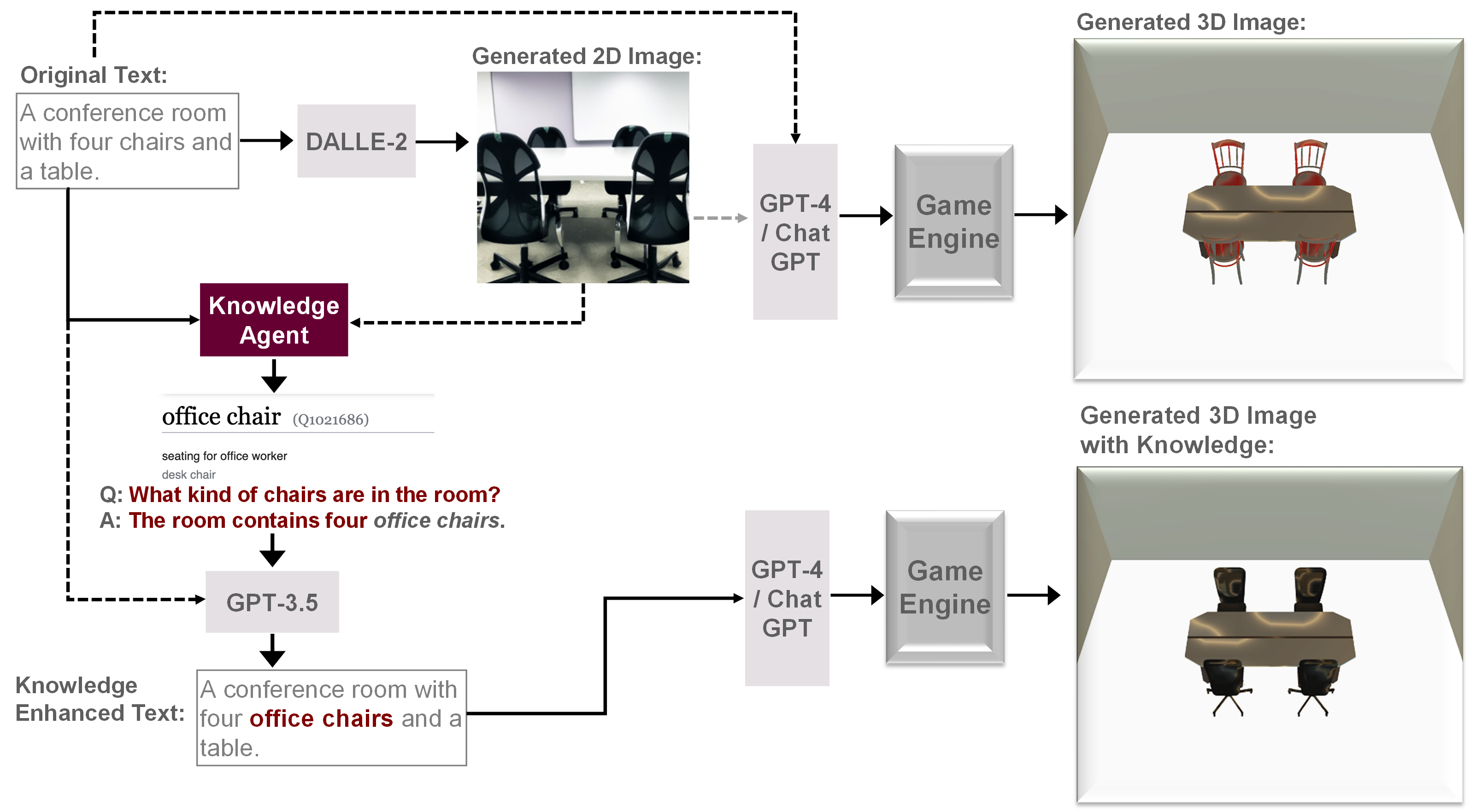} }}%
    \qquad
    \subfloat[\centering]{{\includegraphics[width=0.95\textwidth]{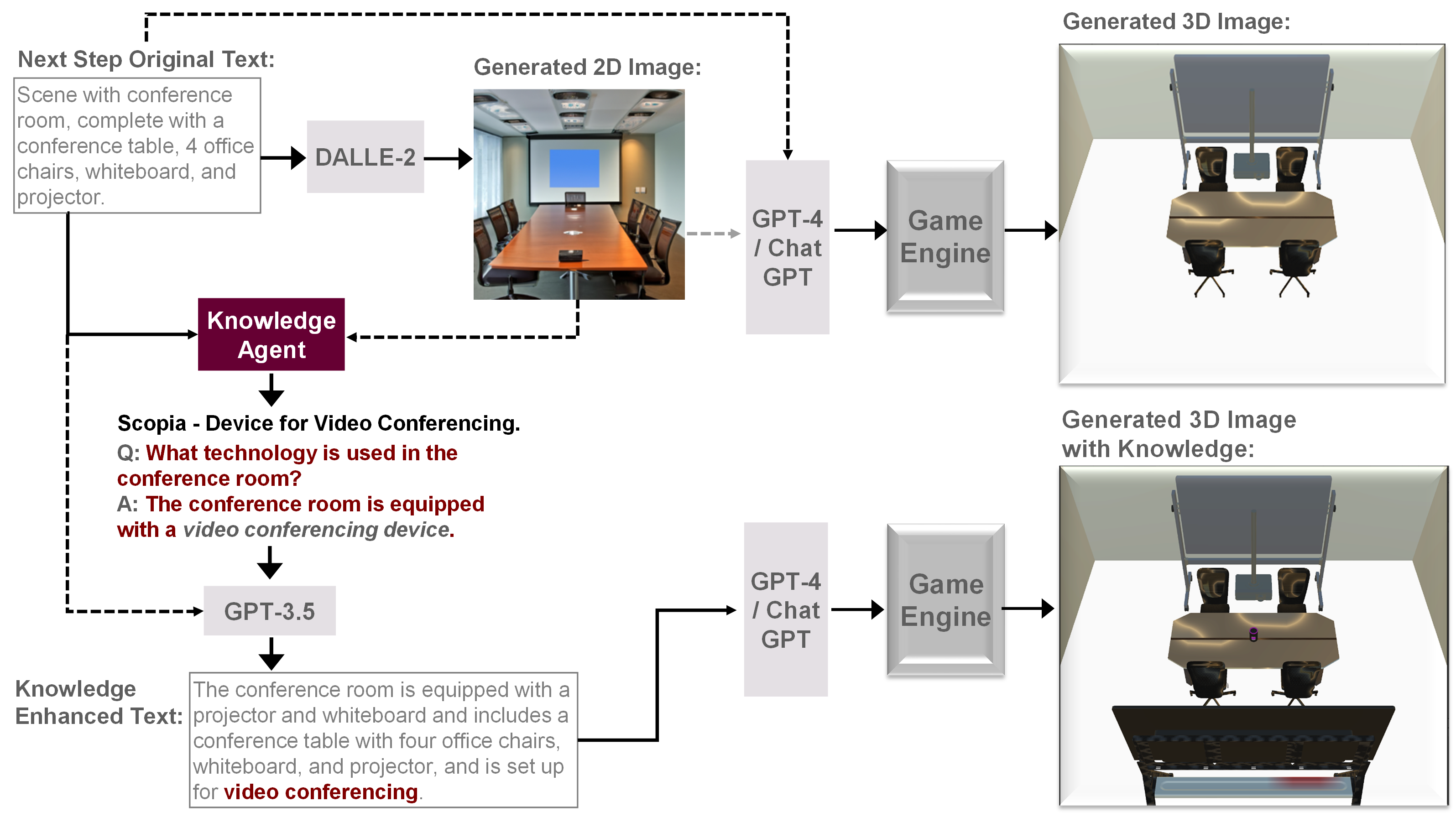} }}%
    \caption{Qualitative examples of 3D scene editing with knowledge enhanced prompts. At inference time, we first generate an image from the input text to learn the prior . The knowledge agent then generates a question and answer tuple which is fed as an input to GPT-3.5. The output of GPT-3.5 is an enhanced version of the input text with added information from external knowledge sources. This text is then given to ChatGPT that outputs the spatial arrangements and low-level program synthesis code. Finally, this code is rendered using Unity engine to output the desired 3D object.}%
    \label{fig:3d_example}%
\end{figure*}

\begin{figure*}[htb!]
\centering
\subfloat[\centering]
{{\includegraphics[width=0.97\textwidth]{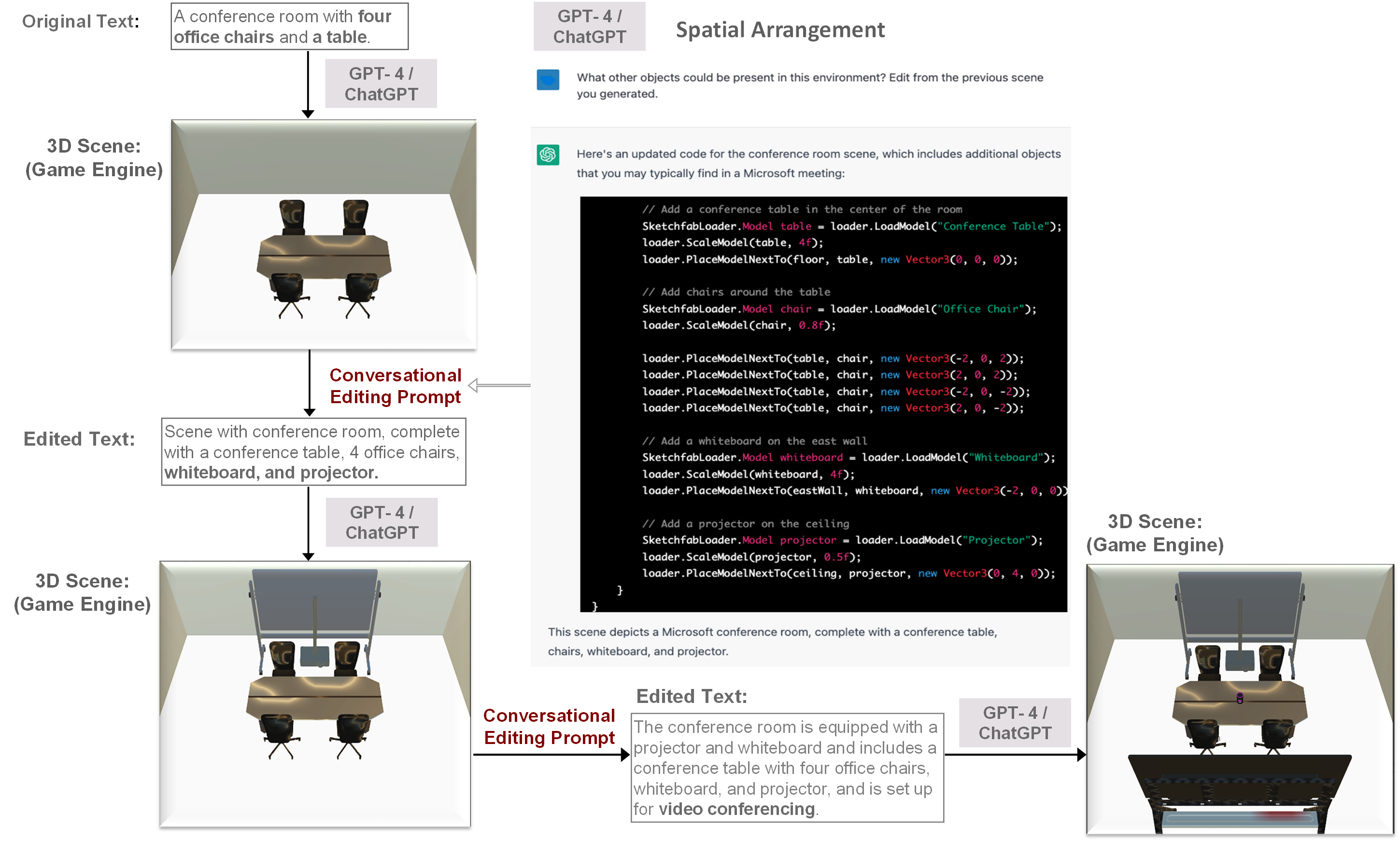} }}
\qquad
    \subfloat[\centering]{{\includegraphics[width=0.95\textwidth]{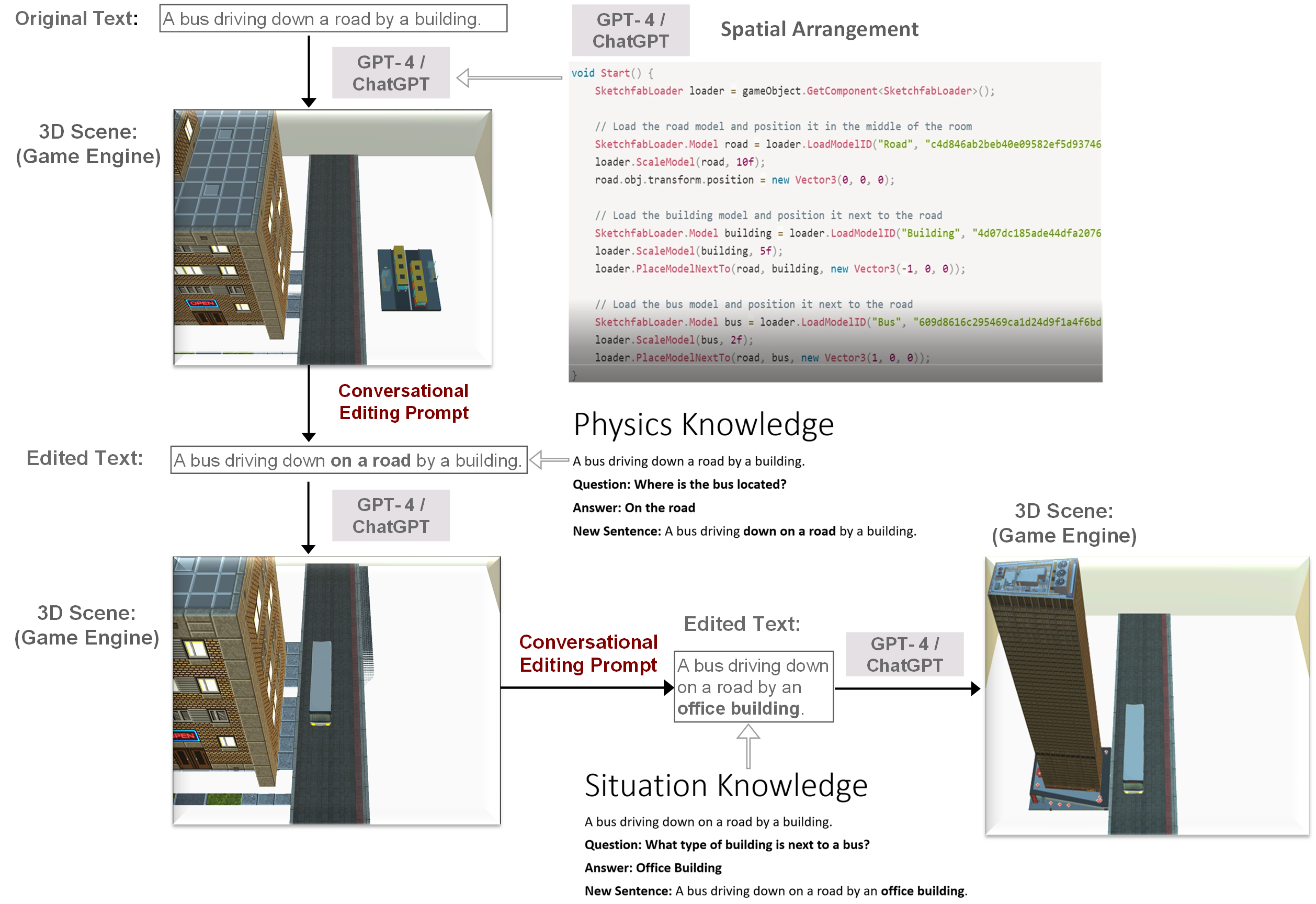} }}
\caption{Editing 3D scene with our trained dialogue interactive scenes projection using GPT4/ChatGPT. We see that GPT4/ChatGPT adds more knowledge context with the spatial arrangement to the original text as shown in the edited text.
}
\label{fig:chatgpt_edit}
\vspace{-0.3cm}
\end{figure*}

\begin{figure*}[htb!]
\centering
    \subfloat[\centering]
{{\includegraphics[width=0.97\textwidth]{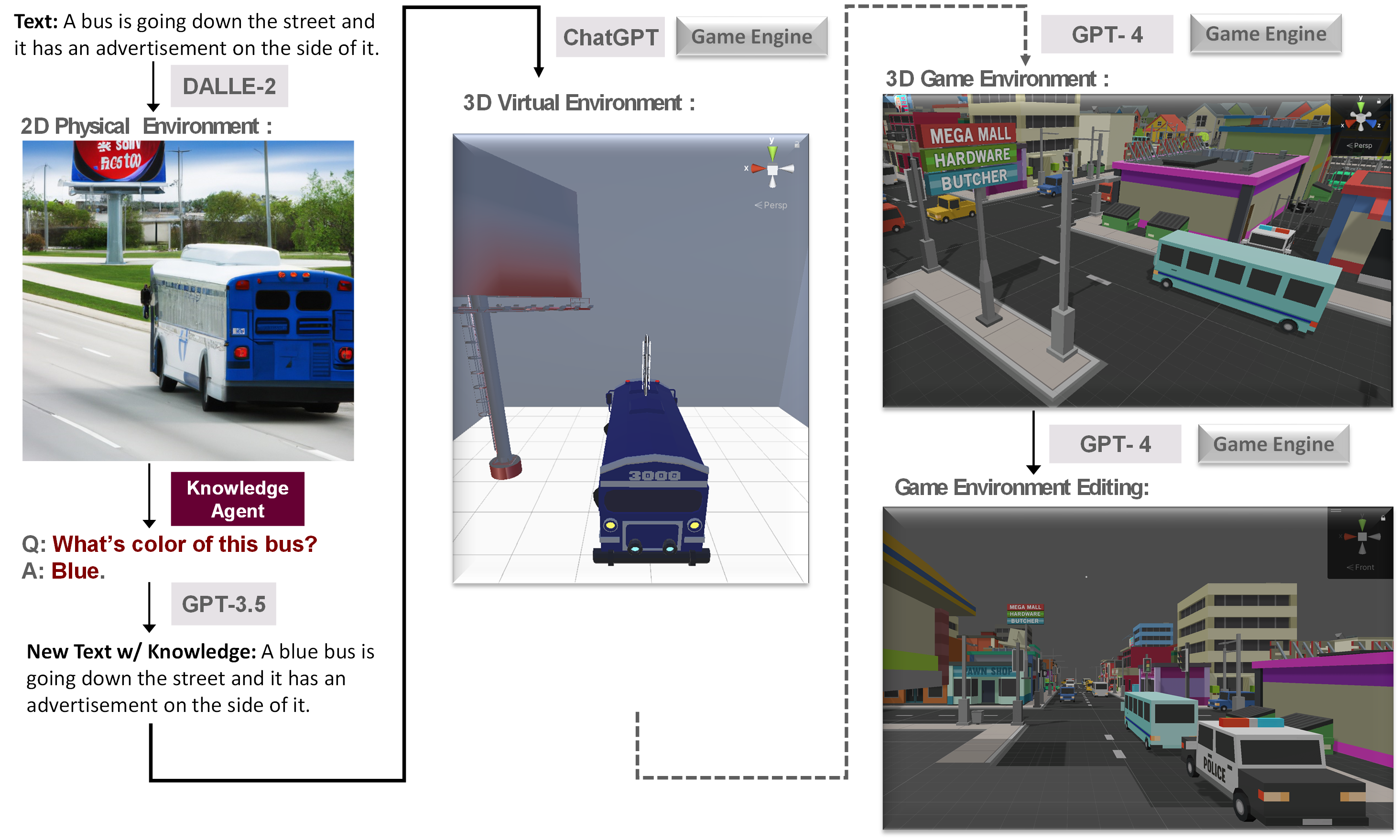}}}
 \qquad
    \subfloat[\centering]{{\includegraphics[width=0.95\textwidth]{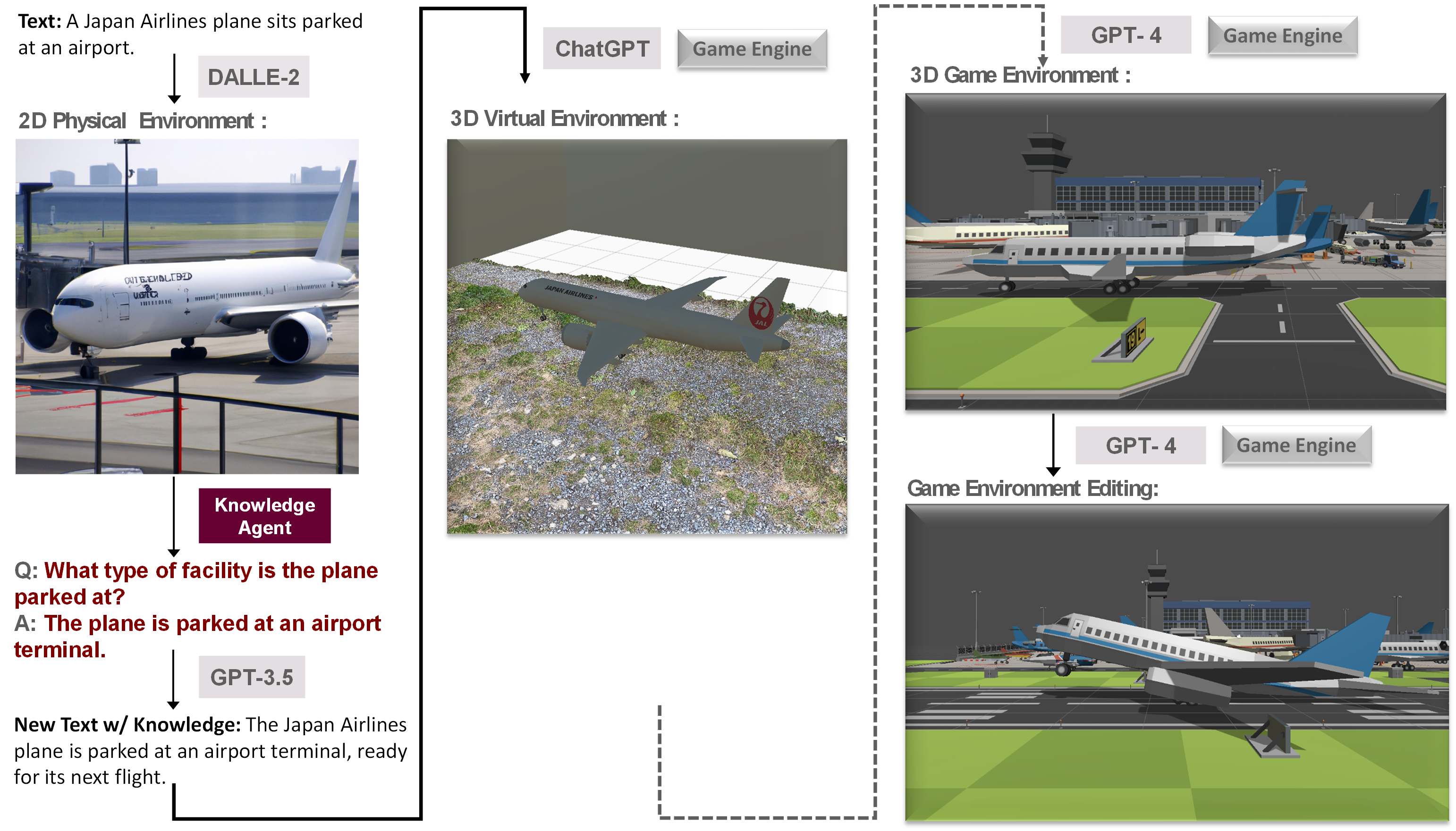} }}%
\caption{Cross-modality and reality-agnostic generation and editing with interactive agent using GPT-4 and ChatGPT.}
\label{fig:gpt4_game}
\vspace{-0.3cm}
\end{figure*}

\subsection{Interactive Cross-modality Generation}
For the interactive cross-modality generation, we retrieve the top 20 knowledge using the embeddings from the CLIP ViT Base-16 model. We use the text captions for the images in validation set of AOK-VQA~\cite{AOKVQA} dataset to perform text to 2D scene generation. To train the question and answer model, we initialize the model with BLIP-large ~\cite{li2022blip} and finetune on the AOK-VQA data in a seq2seq objective with learning rate of $2e^{-6}$, batch size of 128, and for 10 epochs. Policy gradient~\cite{Sutton1999PolicyGM} is used to train the agent after finetuning on AOKVQA data, and the reward is calculated by the CLIP-VIT-base similarity score.
Davinci-003 GPT-3.5 model is used to generate the new knowledge prompts for 2D image, and DALLE-2 images are generated with 256x256 resolution, which we used to generate 2D image and the feedback image for the RL algorithm.

\paragraph{QA-2D image generation and evaluation.}
Figure~\ref{fig:2d_example} shows example of DALLE-2 generated images with original text and knowledge incorporated text query with infinite-memory agent. Please refer more examples of conversational knowledge-2D generation in Appendix~\ref{app:2Dexamples}. We see that by modifying the text query in zero shot setting while keeping the blackbox OpenAI models as frozen, we are able to generate more informative and realistic images. For example, we see more natural portrait of cat drinking water from sink and man looking at himself in the mirror. The last example includes unmentioned entity of GS workstation, but DALLE-2 is able to generate more realistic scene using the entity information. Please find in the table~\ref{tab:human_eval} to find our human-evaluation results of the convectional 2D image generation.



\begin{table*}[t]
\small
    \centering
    \begin{tabular}{lcc}
    \toprule
        {\footnotesize \textbf{Model}} & Relevance (\%) & Naturalness (\%) \\
        \midrule
        \textit{Conversational-2D Scenes Generation} \\
        DALL-E & 78.0 & 81.0 \\
        \textbf{DALL-E w/ Knowledge (ours)} & \textbf{87.0} & \textbf{84.0} \\
        \midrule

        \textit{Conversational-3D Scenes Generation} \\
        GPT4 / ChatGPT - Game Engine & 59.9 & 35.0 \\
        \textbf{GPT4 / ChatGPT - Game Engine w/ Knowledge (ours)} & \textbf{71.1} & \textbf{48.0} \\
    \bottomrule
    \end{tabular}
    \caption{Human evaluation of Conversational-2D (from DALL-E 2) and Conversational-3D scenes generation (from GPT4/ChatGPT and the Game Engine (Unity Stage)). We measure the relevance between scene and text and naturalness of generated scene. We asked yes/no questions to 5 human annotators if the and take the majority vote to get the response on M-Turk\footnote{\url{https://www.mturk.com/}}.}
    \label{tab:human_eval}
    \vspace{-0.3cm}
\end{table*}
\subsection{Virtual Environment Scene Generation, Editing, and Evaluation}



\paragraph{Loading the relevant 3D objects.}
\citet{Roberts2022StepsTP} use the Sketchfab dataset to randomly select 3D models that include the given text input. The caveat with this approach is that the chosen model is not guaranteed to include the correct class of objects due to noisy user labels, as well as their size and orientation. Instead of extracting objects from the Sketchfab dataset, we use the Objaverse~\cite{Deitke2022ObjaverseAU} that provides access to annotated 3D models from the Sketchfab dataset. To ensure that the loaded models correspond to their true objects, we use CLIP~\cite{radford2021learning} to compute a similarity score between the object image and the text and choose the model with the highest score. For each object and text pair, we provide the following prompt: $\texttt{This is an image of \{object\} that}$ $\texttt{refers to \{text\}}$. We observe that providing the entire context helps to retrieve the most relevant object as shown in Figure~\ref{fig:objaverse}.

\paragraph{3D scene generation with knowledge enhanced prompt.}
Figure~\ref{fig:3d_example} shows the pipeline for 3D scene generation using the 2D prior. Without the ground truth image, we only have access to the text query that is used by the agent to retrieve knowledge, we first generate 2D images to determine what the scene would look like in a real world setting. Using the generated image and original text, the agent retrieves explicit knowledge and asks relevant question and answer. Consequently, GPT-3.5 changes the original text to include the knowledge snippet that makes the prompt more informative and realistic in this scene. We use GPT-4/ ChatGPT for the spatial arrangement and the program synthesis generation for 3D scenes with prompt. We next present qualitative results of 3D scene generation with the enhanced knowledge. We finally use GPT4/ChatGPT to generate the knowledge-enhanced prompt of spatial arrangement and programming which is then rendered in the Unity game engine. Note that the OpenAI models (colored in pink) are kept frozen while we only train the agent to perform 3D scene editing. As showed the simulation of the figure~\ref{fig:3d_example}, we observe that with external knowledge the 3D scene a) substitutes wooden chairs to office chairs, and b) includes tools used for video conferencing, making the scene look more realistic.

\paragraph{Cross-modality 3D scene editing.}
Knowledge-enhanced text opens up a novel method of editing 3D scenes with knowledge prior to help improve the naturalness of the scene. We also include the results of editing a 3D scene interactively in a dialogue setting. Figure~\ref{fig:chatgpt_edit} shows an example in which we provide the previously generated the spatial arrangements and low-level program synthesis, and ask GPT4/ChatGPT to fill relevant objects with the new prompt. We observe that GPT4/ChatGPT is able to understand the previous scene and adds relevant environment objects such as whiteboard and projector in the appropriate orientation and location.


\subsection{Reality-agnostic Generation Observation with Emergent Ability.}
We provide DALLE-2 with a text query to generate a real-world 2D image. Then following our pipeline, we give this image as an input to the knowledge agent and consequently the enhanced query is provided to ChatGPT to generate a 3D object loading and the spatial arrangements and the program synthesis for 3D scenes generation. Now this object when further as an input to GPT-4 which adds context to the 3D object by changing the surrounding and appearance of the objects according to the specifications of the scene. Figure~\ref{fig:gpt4_game} showed that examples which we present the VR editing and a novel approach for generating 3D game scene to user shared conversational interactive QA. In contrast to traditional 2D image generation and image-grounded dialogue tasks we simulated on synthesizing 3D-gaming generation and editing content that is relevance and naturalness with the emergent ability of large foundation models for reality-agnostic in generative AI. We show more results for gaming 3D examples in Appendix~\ref{app:msgaming}.


\subsection{Human Evaluation.}
\label{subsec:hmeval}
Since there is no existing metric to auto-evaluate the conversational interactive scene generation, we rely on human evaluation to analyze the results. For each generated scene, we evaluate using scores from 5 humans using crowd-sourcing platform\footnote{\url{https://www.mturk.com/}}. We ask if the generated interactive scene (knowledge-dialogue 2D and 3D) 1) Relevance: matches the text conversational description, and 2) Naturalness: looks realistic. The results are shown in Table~\ref{tab:human_eval}. We see that for both 2D and 3D scene types, knowledge-enhanced text results in more realistic scenes. The 2D scenes greatly benefit from the knowledge in terms of relevance, and both necessarily with the 3D scene generation with the imitation and conversational dialogue way.

\section{Conclusion}
In this work, we explored the infinite knowledge-memory agent to enhance large foundation model for physical and virtual reality scene generation, and found qualitatively that incorporating knowledge in the new prompt itself provides an improvement of generated physical and virtual environments with RL and IL for the foundation models. We investigate the task of knowledge-guided interactive synergistic effects to collaborative scene generation by combining large foundation models, and show promising results of how knowledge agent interactive emergent ability can improve the performance of 2D and 3D scenes generation and editing in our setting. We discussed the emerging capabilities in cross-modality models and reality-agnostic scenarios. It integrates and improves the depth of generalization, conscious and interpretability of a complex adaptive AI systems. Emergent ability works in Generative AI for metaverse and gaming simulation. We leave it as future work to explore real-world different types of human knowledge feedback and thorough investigation of more learning algorithms to improve our pipeline (e.g., prosody, anaphora, gesture, etc.) for the mix-reality generative AI.

\section*{Ethics Statement}

LLM has many applications. In addition to 2D and 3D generation, grounded language models could help drive content generation and editing for bots and AI agents, and assist in productivity applications, helping to re-write, paraphrase, translate or synthesize text. Fundamental advances in text derived 2D and 3D generation help contribute towards these goals and many would benefit from a greater understanding of how to model emergent ability and empathetic with language and image in the physical world. Arguably many of these applications could have positive benefits. 

However, the emerging ability technology could also be used by bad actors. AI systems that generate content can be used to manipulate or deceive people. Therefore, it is very important that this technology is developed in accordance with responsible AI guidelines. For example, explicitly communicating to users that content is generated by an AI system and providing the user with controls in order to customize such a system. It is possible the emerging ability could be used to develop new methods to detect manipulative content - partly because it is rich with robotic empathy with LLM and virtual environment generation - and thus help address another real world problem. 

\section*{Acknowledgement}

We are especially grateful to Antonio Criminisi, Tom Cashman, Andy Wilson, Bala Kumaravel, Andrzej Banburski-Fahey, Corby Rosset, Michel Galley, Jianwei Yang, Pan Lu, Mary Czerwinski, Kiran Muthabatulla, Silviu Cucerzan, Ahmed Awadallah, Saleema Amershi, Nguyen Bach, Jennifer Marsman, Jaron Lanier for their comments, suggestions, painstaking multiple reviews of this paper, and their pointers to the literature.

\bibliography{example_paper}
\bibliographystyle{acl_natbib}

\clearpage



\appendix




\section*{Appendix}

\section{Knowledge-Tensor CLIP Pre-Training Details}
\label{sec:app-tensor-clip}
The Knowledge-Tensor CLIP model is trained to align the image, text, and knowledge modalities together to optimize the knowledge retrieval mechanism using the both image and text modalities. Inspired by the effectiveness of masked training in BEIT-3 \cite{beit3}, we add the decoder-based masked loss in the visual and text encoders, in which the masked image patches and masked text tokens are given as input. The contrastive learning is applied to the same masked inputs during training to allow the model to reason over different image regions and text. The masking loss is not applied to the knowledge encoder the knowledge embeddings are kept as frozen throughout training. We randomly mask 15\% tokens of texts and mask 40\% of image patches using a block-wise masking strategy as in BEIT2 \cite{beitv2}. The effect of masked loss is shown in Table~\ref{tab:knowledge_retrieval}, which provides a slight boost in the downstream task evaluations.



\begin{figure*}
\centering
\includegraphics[width=0.9\textwidth]{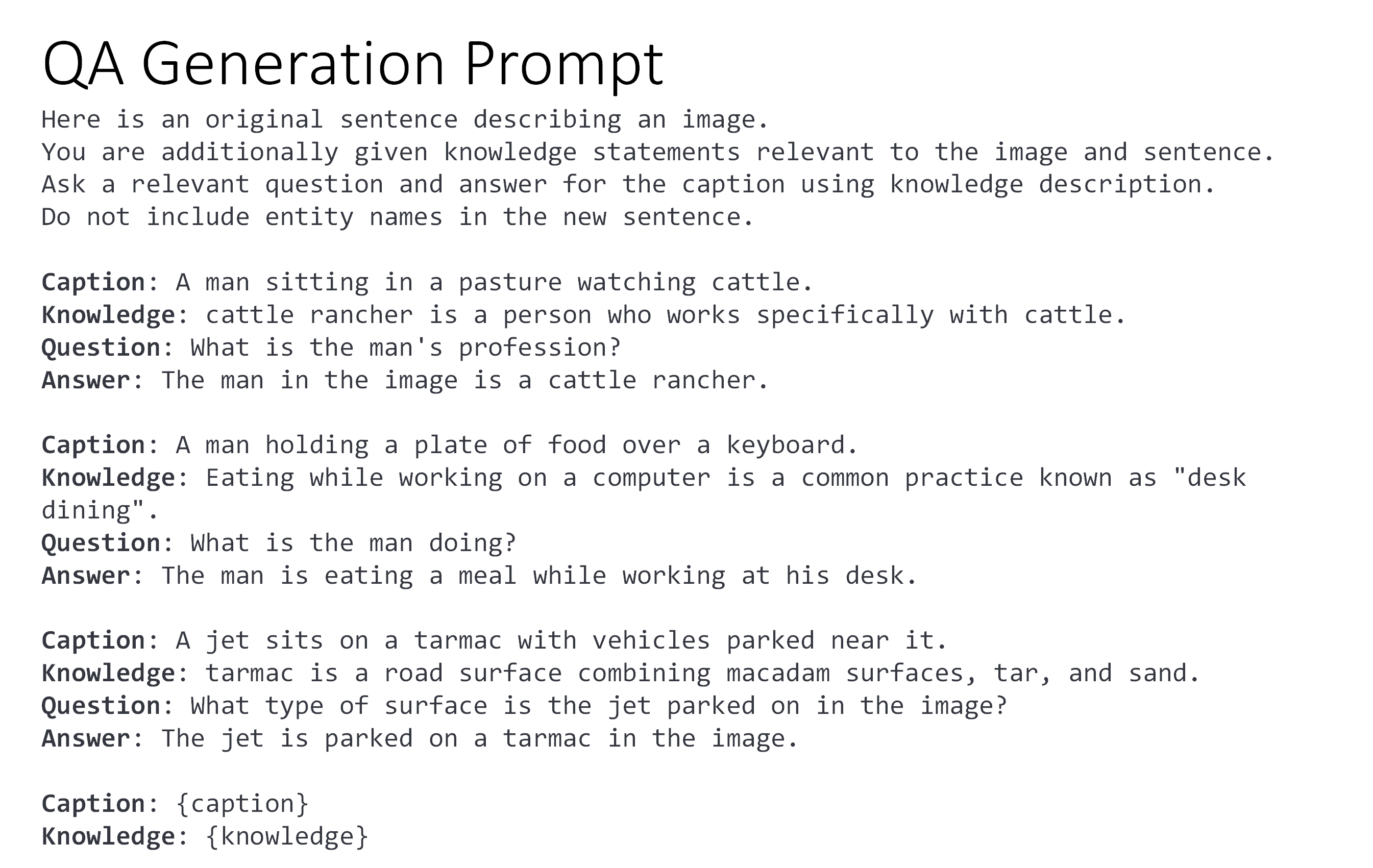}
\caption{An example of the evidence of rationale QA that we obtain from GPT-3.5 by using a combination of image and text candidate to query it.}
\label{fig:prompt1}
\vspace{-0.3cm}
\end{figure*}


\begin{figure*}[h]
\centering
\includegraphics[width=0.96\textwidth]
{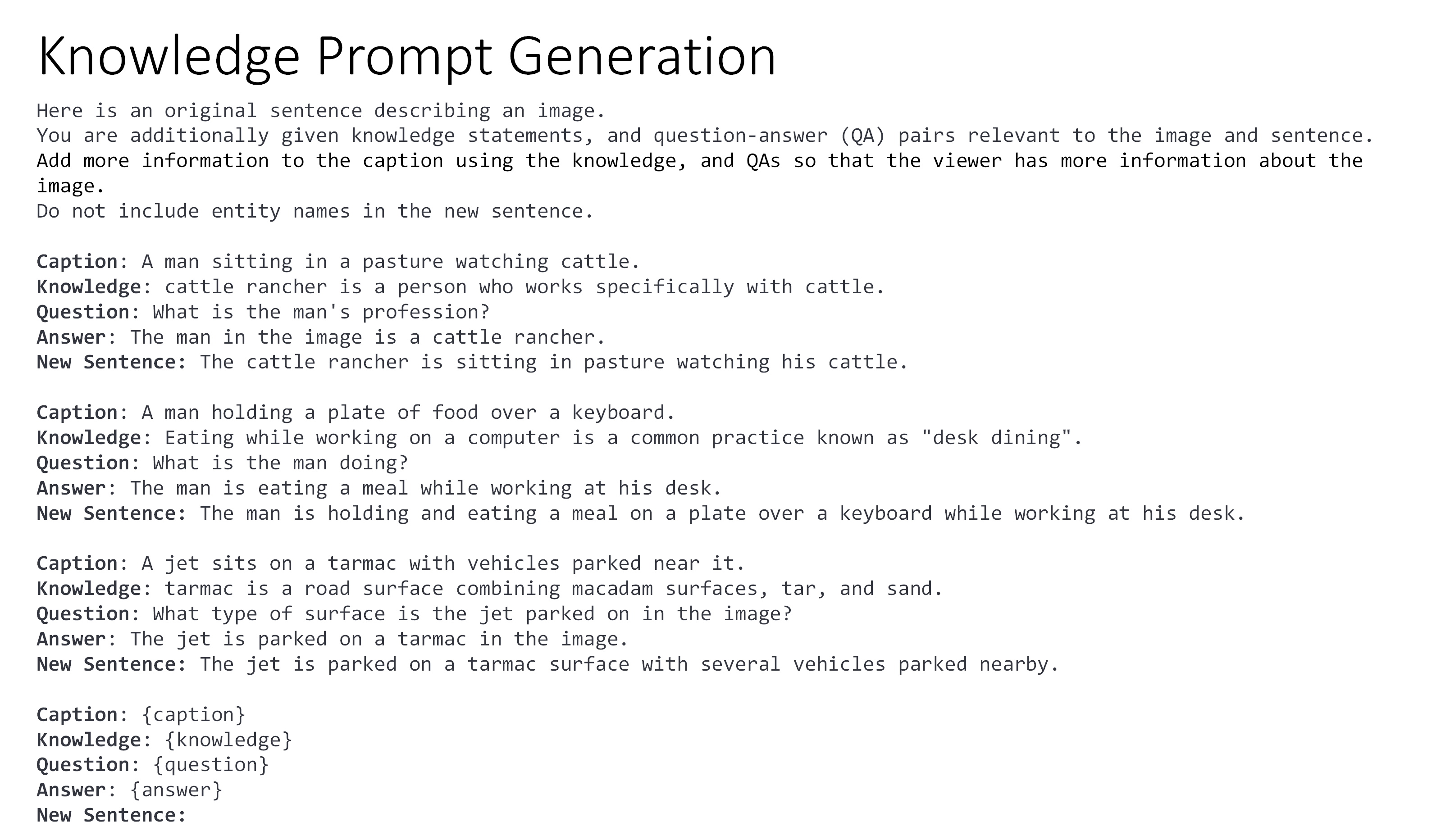}
\caption{An example of the prompts that we use to query GPT-3.5 in our knowledge-augmented GPT-3.5 query system.
}
\label{fig:prompt2}
\vspace{-0.3cm}
\end{figure*}
\section{More examples for conversational knowledge 2D image generation Examples. }
\label{app:2Dexamples}
We show more examples for generating the 2D images using the infinite knowledge-memory agent in Table~\ref{tbl:examples}.

\section{Effect of knowledge on losses}
\label{app:analyzation}
\begin{figure*}[h]
\centering
\includegraphics[width=\columnwidth]
{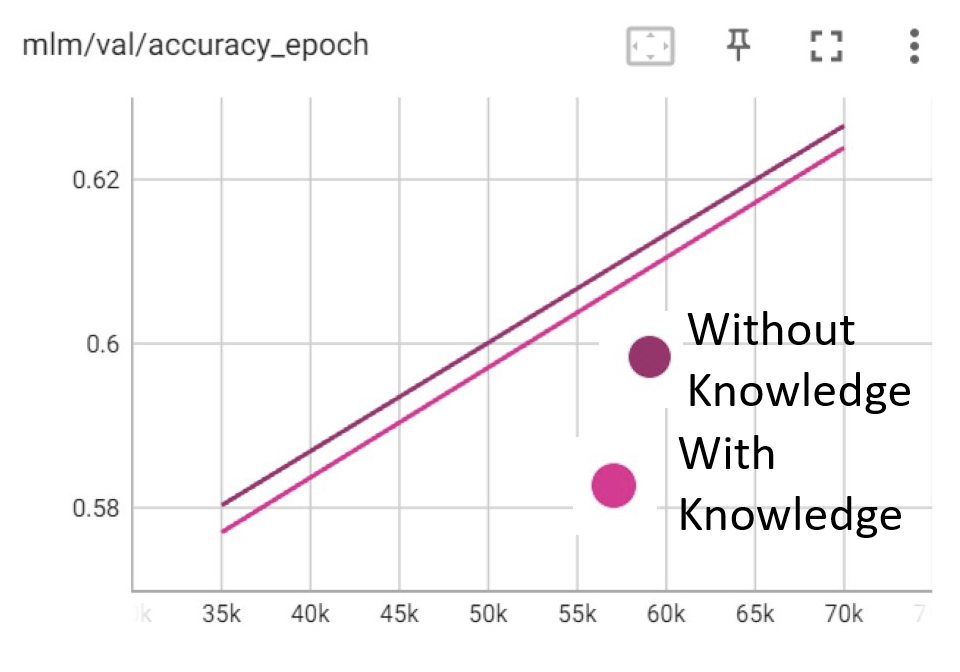}
\includegraphics[width=\columnwidth]
{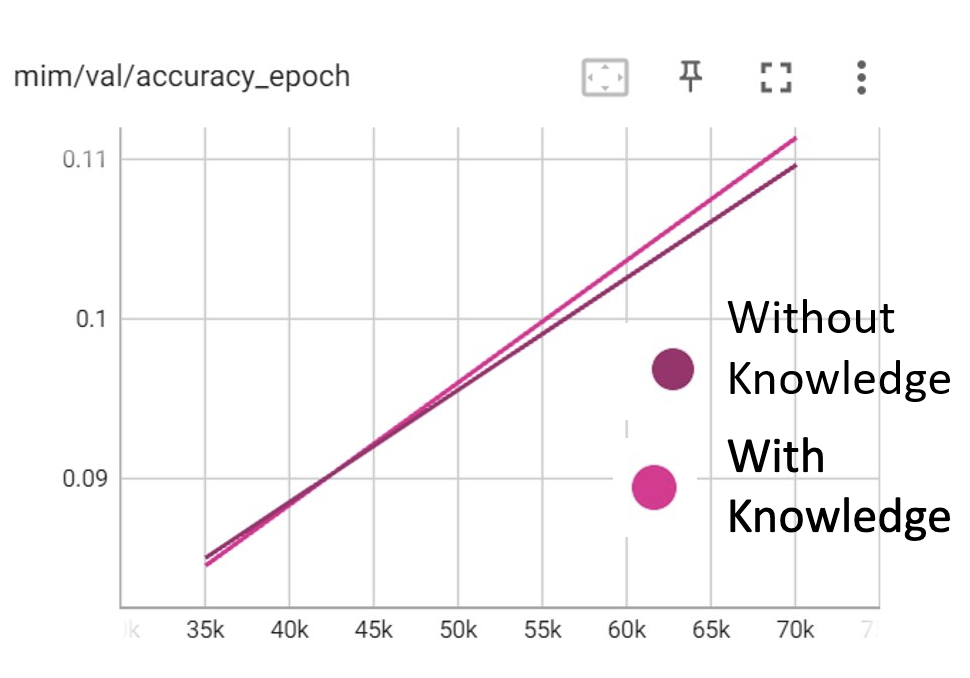}
\caption{Effect of knowledge on both language and image losses.}
\label{fig:loss_cmp}
\vspace{-0.3cm}
\end{figure*}

We analyze the effect of knowledge on both types of loss functions i.e. language loss and image loss. The results are shown in Fig~\ref{fig:loss_cmp}. We observe that knowledge helps in improving the image loss much better than the language loss. This can be attributed to knowledge providing better learning signal for 2D image generation and hence results in producing more realistic images.

\section{Prompt for Knowledge-Memory Agent}

\subsection{Knowledge-based Question Answer Generation}
The prompt for GPT-3.5 to provide additional question-answering supervision data for the knowledge memory agent is shown in Figure~\ref{fig:prompt1}. We retrieve the top $K$ knowledge using the Tensor-CLIP model with the COCO image and text captions. With knowledge $K$ as context, question and answer are generated to train the agent.

\subsection{Prompt for Knowledge-Enhanced Query Generation}
Figure \ref{fig:prompt2} shows a prompt to query GPT-3.5 that utilizes the retrieved knowledge, question and answer generated by the agent to reformulate the original text into a `knowledge-enhanced" description.



\subsection{Prompt for VR and Game Scene Generation for GPT Models}

\begin{figure*}[h]
\centering
\includegraphics[width=0.97\textwidth]
{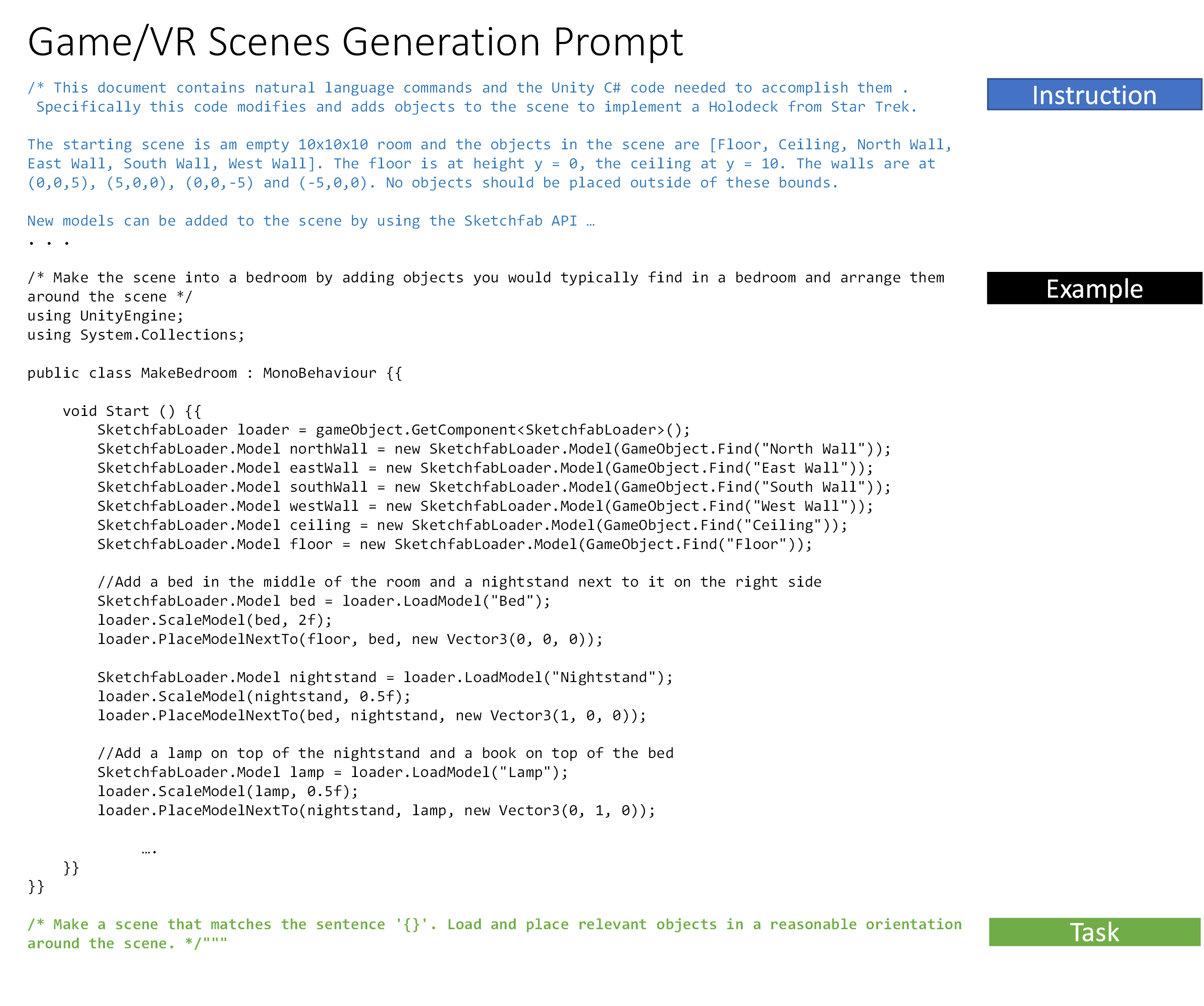}
\caption{An example of the prompts for VR and game scene generation and editing which we use GPT4/ChatGPT in our knowledge-augmented query system.
}
\label{fig:prompt3}
\vspace{-0.3cm}
\end{figure*}

In Figure \ref{fig:prompt3}, we provide ChatGPT with a text prompt to generate the program synthesis that is then rendered in the Unity game engine. If the query is about generating a scene in a game, the model is able to find the relevant context information, such as size, physics, relative orientations, and other relative objects in the environment, and output a 3D scene that resembles the scenes from the game. It uses the Sketchfab assets that could be easily placed in the Unity game engine.

\section{Human Evaluation Details}
\begin{figure*}[h]
\centering
\includegraphics[width=0.97\textwidth]
{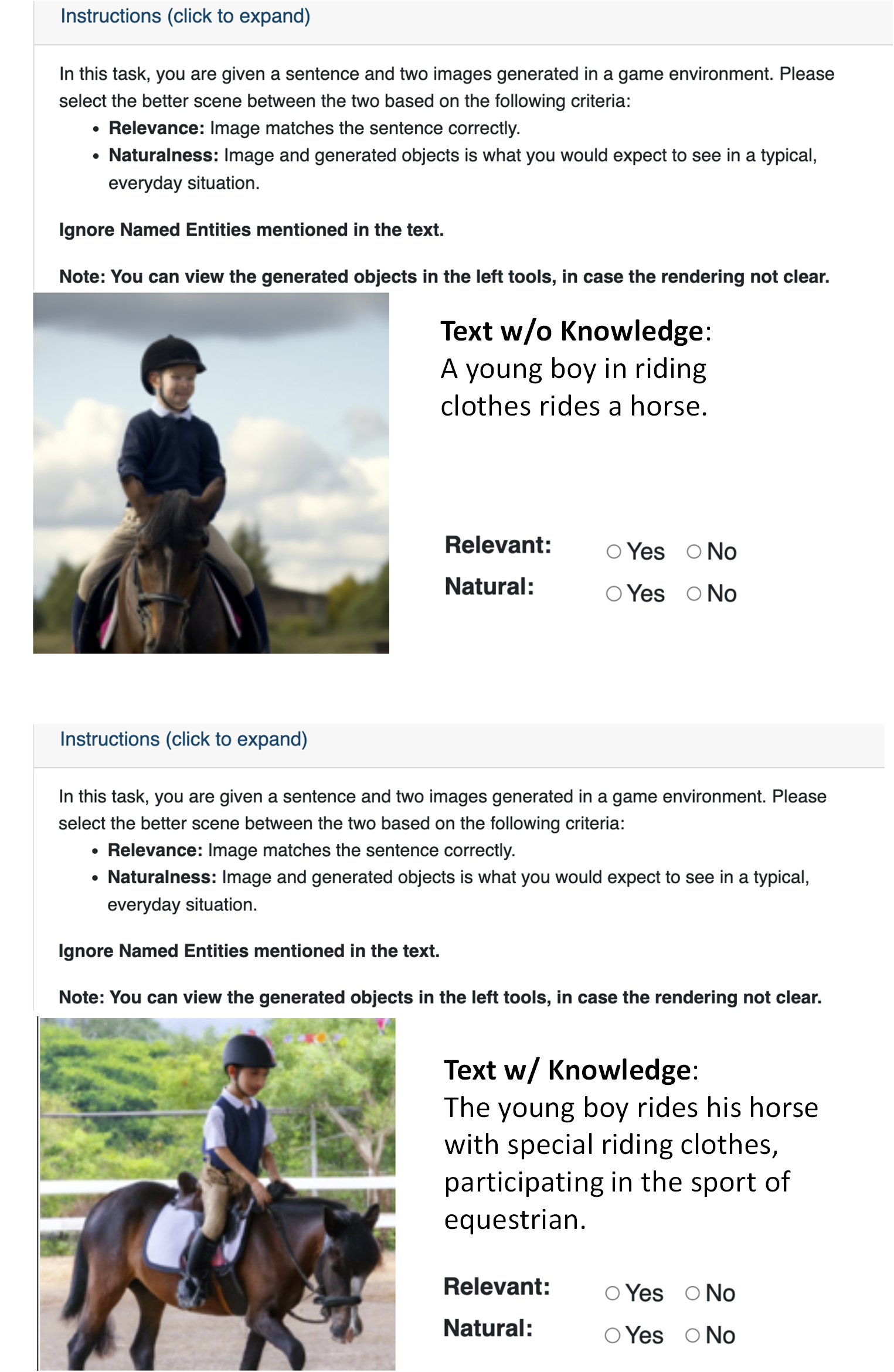}
\caption{An example of the conversational 2D human evaluation.
}
\label{fig:2DHE}
\vspace{-0.3cm}
\end{figure*}

In Fig.~\ref{fig:2DHE}, we show screenshots of the instructions that were given to the participants for the human evaluation study using mechanical turk. The results shown in Sec~\ref{subsec:hmeval} are based on two metrics shown in the figure here: Relevance and Naturalness. The users have to select which scene is more relevant and natural separately for both 2D and 3D images. Three human evaluators are asked to choose binary yes/no choice for each image, and we take the average of answers to evaluate the scene generation performances. 

\section{Microsoft Gaming Scenario}
\label{app:msgaming}
\begin{figure*}[htb]
\centering
    \subfloat[\centering]
{{\includegraphics[width=0.97\textwidth]{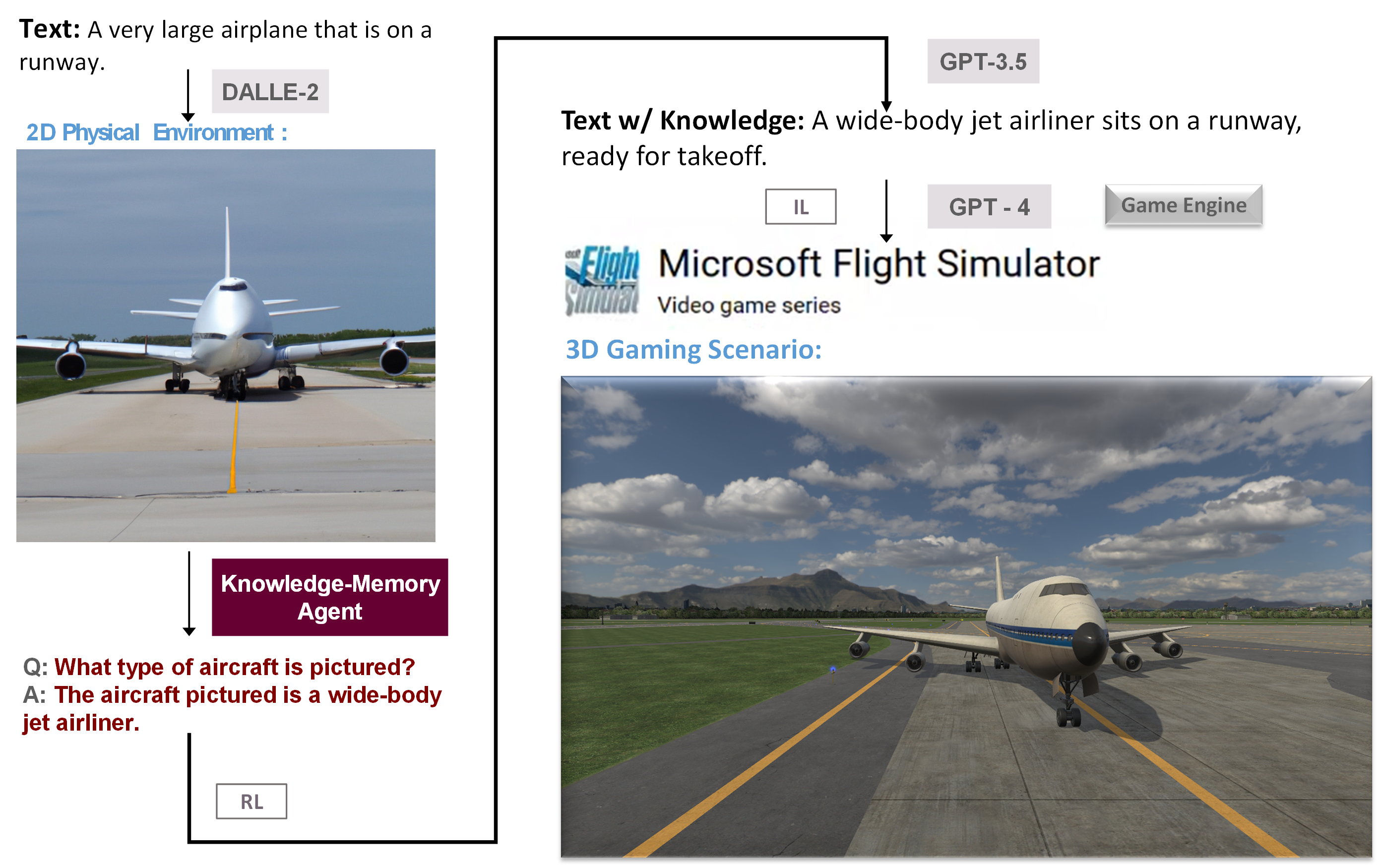}}}
 \qquad
    \subfloat[\centering]{{\includegraphics[width=0.95\textwidth]{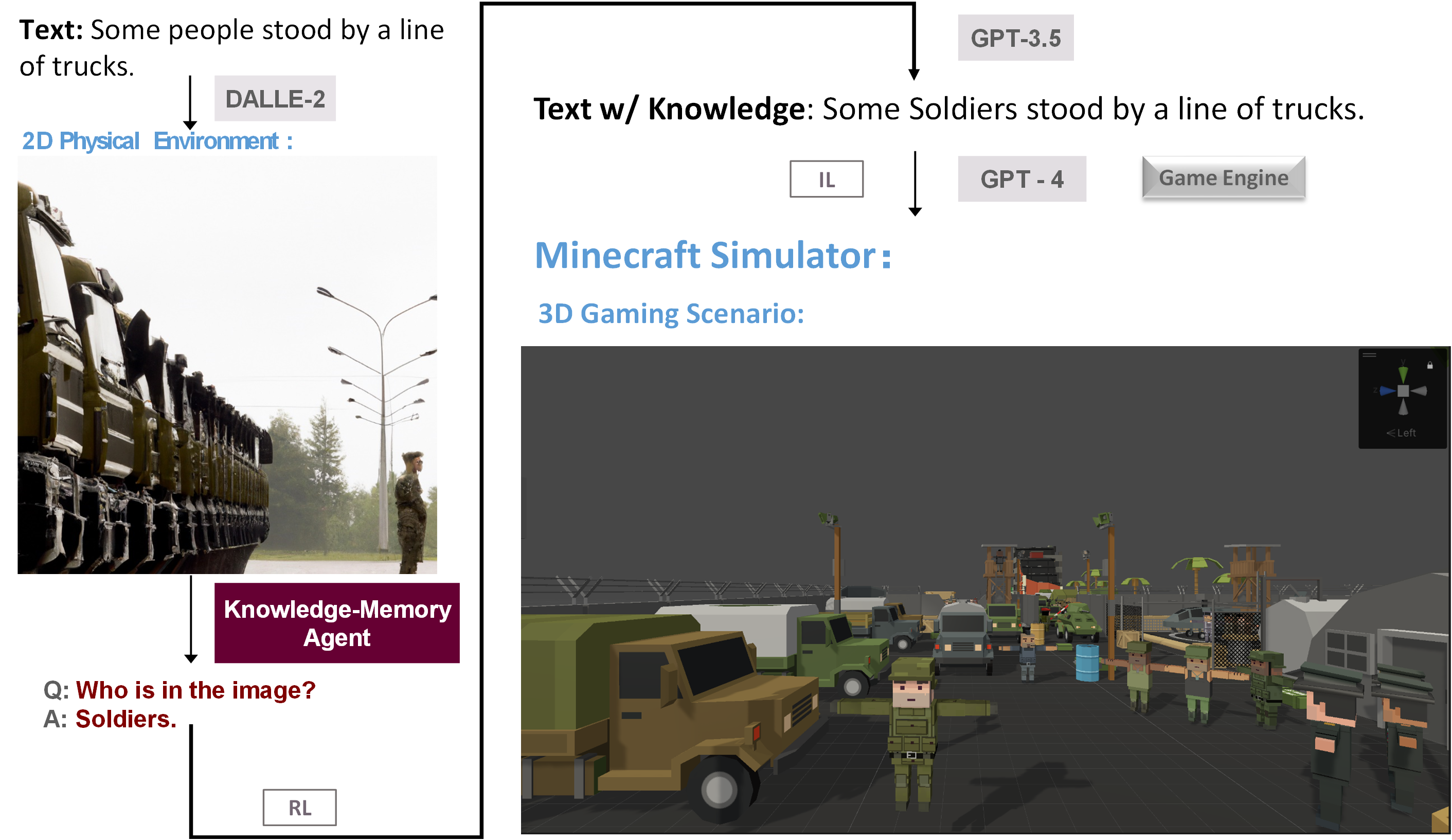} }}%
\caption{Two examples of Microsoft 3D Game Scenario. One of the knowledge interactive simulation in Microsoft Fight Simulator; another is knowledge interactive simulation in Minecraft scene generator.}
\label{fig:MSgame}
\vspace{-0.3cm}
\end{figure*}

We showed the examples of Microsoft 3D Game Scenario. One of the knowledge interactive simulation in Microsoft Fight Simulator; another is knowledge interactive simulation in Minecraft scene generator. The details please refer the Fig.~\ref{fig:MSgame}.


\clearpage
\begin{table*}[h!]
  \centering
  \begin{adjustbox}{width=\textwidth}
  \begin{tabular}{ | m{3cm} | m{3cm} | m{3cm} | m{3cm} | c | }
    \hline
    Text & Knowledge & Q,A & New Prompt & Image \\ \hline
    Cat sleeping in front of a powered on laptop computer.
    &
    computer is a general-purpose device for performing arithmetic or logical operations
    &
    \textbf{Q:} What is the device in the image?
    \textbf{A:} The device in the image is a laptop computer.
    &
    The cat is peacefully sleeping in front of a powered on laptop computer.	
    & 
    \begin{minipage}{.3\textwidth}
      \includegraphics[width=\linewidth, height=40mm]{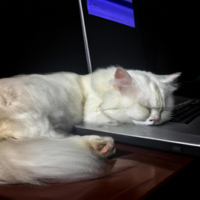}
    \end{minipage}
    \\ \hline
    A group of people riding skis on a snowy surface
    &
    skiing as transport or recreation is a the use of (snow)skis for transport purposes or as recreation	
    &
    \textbf{Q:} What activity is the group participating in?	
    \textbf{A:} The group is skiing as a form of transport or recreation. 
    &
    The group is skiing on the snow-covered surface as a form of transport or recreation.
    &
    \begin{minipage}{.3\textwidth}
      \includegraphics[width=\linewidth, height=40mm]{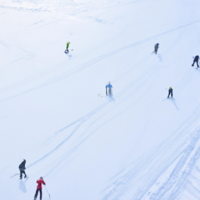}
    \end{minipage}
    \\ \hline
    A woman is holding a slice of red and white cake	
    &
    sweetheart cake is a food	
    &
    \textbf{Q:} What type of food is the woman holding?	
    \textbf{A:} The woman is holding a slice of sweetheart cake.
    &
    The woman holds a slice of sweetheart cake, decorated in red and white.	
    &
    \begin{minipage}{.3\textwidth}
      \includegraphics[width=\linewidth, height=40mm]{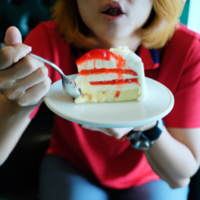}
    \end{minipage}
    \\ \hline
    A person jumping up in the air on a skateboard.	
    &
    street skateboarding is a style of skateboarding	
    &
    \textbf{Q:} What type of skateboarding is the person doing?	
    \textbf{A:} The person is doing street skateboarding.	
    &
    The person is doing a street skateboarding trick, jumping up in the air on a skateboard.	
    &
    \begin{minipage}{.3\textwidth}
      \includegraphics[width=\linewidth, height=40mm]{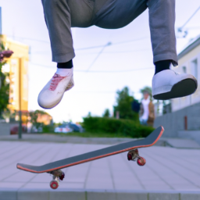}
    \end{minipage}
    \\ \hline
  \end{tabular}
  \end{adjustbox}
  \caption{Knowledge based 2D image generation examples in physical world.}\label{tbl:examples}
\end{table*}

\begin{figure*}[h]
\centering
\includegraphics[width=0.55\textwidth]
{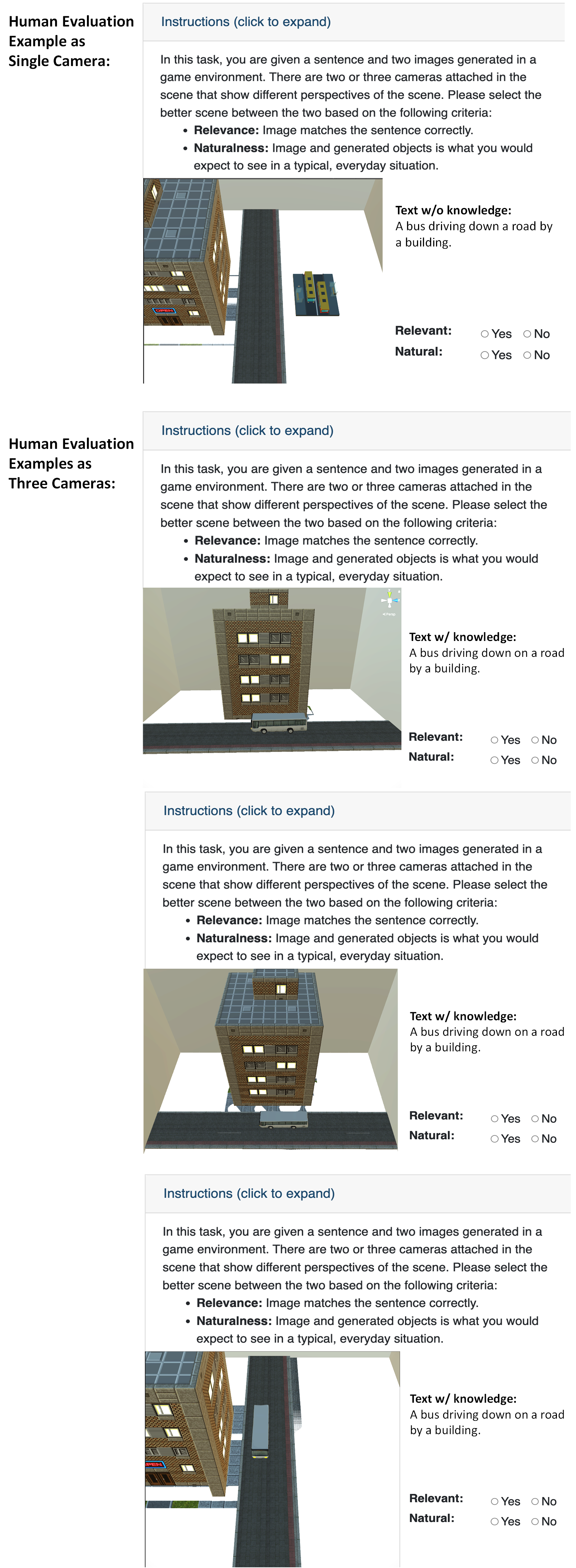}
\caption{Two examples of human evaluation for the conversational 3D VR Scenario. One is for single camera example; another is for the three cameras examples.
}
\label{fig:3DHumanE}
\vspace{-0.3cm}
\end{figure*}


\end{document}